\documentclass[preprint,12pt]{elsarticle}
\usepackage{amsmath,amsfonts}
\usepackage{algorithmic}
\usepackage{algorithm}
\usepackage{array}
\usepackage[caption=false,font=normalsize,labelfont=sf,textfont=sf]{subfig}
\usepackage{textcomp}
\usepackage{stfloats}
\usepackage{url}
\usepackage{verbatim}
\usepackage{graphicx}
\usepackage{cite}
\usepackage[utf8]{inputenc} 

\usepackage{balance}
\usepackage{amsmath,amssymb}
\usepackage{url}

\usepackage{color}
\usepackage{xcolor}

\usepackage{xspace}
\usepackage{graphicx}
\usepackage{longtable}
\usepackage{caption}
\usepackage{amsmath,amsthm}
\usepackage{multirow}
\usepackage{multicol}
\usepackage{booktabs}
\usepackage{url}
\usepackage{pifont}
\usepackage{color}
\usepackage{supertabular}
\usepackage[export]{adjustbox}

\usepackage{amsmath}
\usepackage{amsthm,amsmath,amssymb}
\usepackage{mathrsfs}
\usepackage{bm}
\usepackage{hyperref}

\usepackage{CJKutf8}

\usepackage{cleveref}

\floatname{algorithm}{Algorithm}

\newcommand{\vct}[1]{\ensuremath{\boldsymbol{#1}}}
\newcommand{\mat}[1]{\ensuremath{\mathbf{#1}}}

\newcommand{\argmax}{\operatornamewithlimits{\arg\,\max}}

\newcommand{\argmin}{\operatornamewithlimits{\arg\,\min}}

\newcommand{\myparagraph}[1]{\smallskip \noindent \textbf{#1}}
\newcommand{\ie}{{i.e.}\xspace}
\newcommand{\eg}{{e.g.}\xspace}

\newcommand{\yzzedit}[1]{\textcolor{black}{#1}}
\newcommand{\aedit}[1]{\textcolor{black}{#1}}

\newcommand{\tocheck}[1]{\textcolor{black}{#1}} %

\begin{document}

\begin{frontmatter}

\author[NPU]{Yang Zheng}

\author[unige]{Luca Demetrio}

\author[cispa]{Antonio Emanuele Cinà}

\author[NPU]{Xiaoyi Feng}

\author[NPU]{Zhaoqiang Xia}

\author[NPU]{Xiaoyue Jiang}

\author[unica]{Ambra Demontis\corref{mycorrespondingauthor}}
\cortext[mycorrespondingauthor]{Corresponding author}
\ead{ambra.demontis@unica.it}

\author[unica]{Battista Biggio}

\author[NPU,unige]{Fabio Roli}

\address[NPU]{Northwestern Polytechnical University, Xi’an, China}
\address[unige]{University of Genoa, Italy}
\address[unica]{University of Cagliari, Italy}
\address[cispa]{CISPA Helmholtz Center for Information Security, Germany}

\title{Hardening RGB-D Object Recognition Systems against Adversarial Patch Attacks}



\begin{abstract}
RGB-D object recognition systems improve their predictive performances by fusing color and depth information, outperforming neural network architectures that rely solely on colors.
While RGB-D systems are expected to be more robust to adversarial examples than RGB-only systems, they have also been proven to be highly vulnerable. Their robustness is similar even when the adversarial examples are generated by altering only the original images' colors.
Different works highlighted the vulnerability of RGB-D systems; however, there is a lacking of technical explanations for this weakness. Hence, in our work, we bridge this gap by investigating the learned deep representation of RGB-D systems, discovering that color features make the function learned by the network more complex and, thus, more sensitive to small perturbations.
To mitigate this problem, we propose a defense based on a detection mechanism that makes RGB-D systems more robust against adversarial examples.
We empirically show that this defense improves the performances of RGB-D systems against adversarial examples even when they are computed ad-hoc to circumvent this detection mechanism, and that is also more effective than adversarial training.
\end{abstract}


\end{frontmatter}
\section{Introduction}

Object classification systems are machine learning models that classify objects depicted inside input photos. 
The acquisition of pictures destroys the information about the depth since images are projections of 3D objects on a flat 2D rectangular surface, hence losing meaningful information in the process.
To overcome this loss, RGB-D systems fuse the information acquired through regular RGB cameras with the depth information retrieved with specific sensors and techniques.
Such multi-modality is more reliable than the information provided by color alone and enables more accurate classification thanks to the additional knowledge retrieved from the spatial properties of the objects~\citep{loghmani2019recurrent}.
Even if it might be reasonable to think that the addition of the depth information could lead to a more robust system, previous work~\citep{GeirhosRMBWB19, abdelfattah2021adversarial, yu2020investigating} have shown that RGB-D models are vulnerable, as well as RGB systems, against \emph{adversarial examples}~\citep{biggio2013evasion,szegedy2014properties}: minimally-perturbed samples that cause the target model to misbehave at test time. 
In particular, they highlighted that attackers that can manipulate both RGB and depth features have the complete control over the detection capability of the target system.
However, even if these attacks were successful, it is difficult to understand is such vulnerability is principally caused by one family of features alone.
For instance, Yu et al.~\citep{yu2020investigating} indicate that different strategies are more or less effective on the RGB or depth, alternatively, without drawing any conclusion on the matter.
Also, Geirhos et al.~\citep{GeirhosRMBWB19} study the efficacy of spatial and RGB features, concluding that shape and depth can help machine learning models to increase their predictive accuracy and that both are subject to attacks, without investigating the latter.
Thus, it is established that the fusion of both RGB and depth information grants machine learning models the ability to better recognize objects, but such a discussion is completely missing regarding the cause of their weaknesses.

In this work, we bridge this gap by analyzing why adversarial attacks against RGB-D systems are effective. 
To evaluate their performances in realistic conditions, we assess their robustness also against \textit{adversarial patches}, contiguous chunks of pixel values optimized to produce misclassifications, which can be easily applied physically on images and objects as printed stickers~\citep{brown2017adversarial}.
Although RGB-D systems consider both the colors and the depth, an attacker can easily subvert their performances by optimizing attacks targeting the sole RGB layer.
To explain this phenomenon, we investigate how the internal layers of neural networks transform data during the forward pass.
To this end, we compute pair-wise distances between each layer of models trained with RGB or depth information using the Centroid Kernel Alignment (CKA), and we highlight that RGB induces higher variability than the depth channel.
On the other hand, the internal representations learned at training time using only the depth information are similar considering the pair-wise distances of the layers, producing a smoother decision function that is more difficult to exploit by adversarial attacks.

Then, we show how a defense based on detection can be used to reduce the vulnerability of RGB-D systems.
Each input sample is processed to obtain its RGB-D representation and compared with the predicted class's centroid.
The input is classified as an adversarial example if these two differ more than a threshold.
We show that our defense effectively increases the robustness of the victim model not only against adversarial examples unaware of the defense in place but also against adaptive attacks aware of the detection mechanism and designed to overcome it.
Lastly, we test the efficacy of the detector compared to an adversarially-trained model~\citep{wang2022adversarial}, that base its robustness on the inclusion of adversarial examples at training time.
Our results suggest that our detector better handles the presence of attacks, by keeping good performance against attackers with increasing strenght.

Thus, the main contributions of this work can be summarized as follows: 
\begin{itemize}
    \item we empirically assess the performance of a state-of-art object recognition system based on both RGB and depth features, by considering different neural networks as backbone for the considered fusion model;
    \item we explain why RGB features are less robust than depth features, by measuring the variability learned at training time by each internal layer of the analyzed model;
    \item we develop a defense that detects out-of-distribution samples, and we compare its performance with adversarial training, the only defense proposed to secure RGBD systems, showing that our methodology achieves better robustness.
\end{itemize}
The rest of the paper is organized as follows.
We first introduce the background concepts needed to understand the RGB-D systems and the threats posed by adversarial examples (\autoref{sec:background}).
We continue by discussing a methodology to interpret and understand the robustness of RGB-D models (\autoref{sec:robustness}), and we explain how these systems can be defended (\autoref{sec:defense}).
We follow by discussing our empirical findings (Section~\ref{sec:experiments_results}). We conclude our paper by offering an overview of the related work (\autoref{sec:related}) and by discussing limitations and future work of our study (\autoref{sec:conclusions}).

\begin{figure*}[t]
    \centering
    \includegraphics[width=0.99\textwidth]{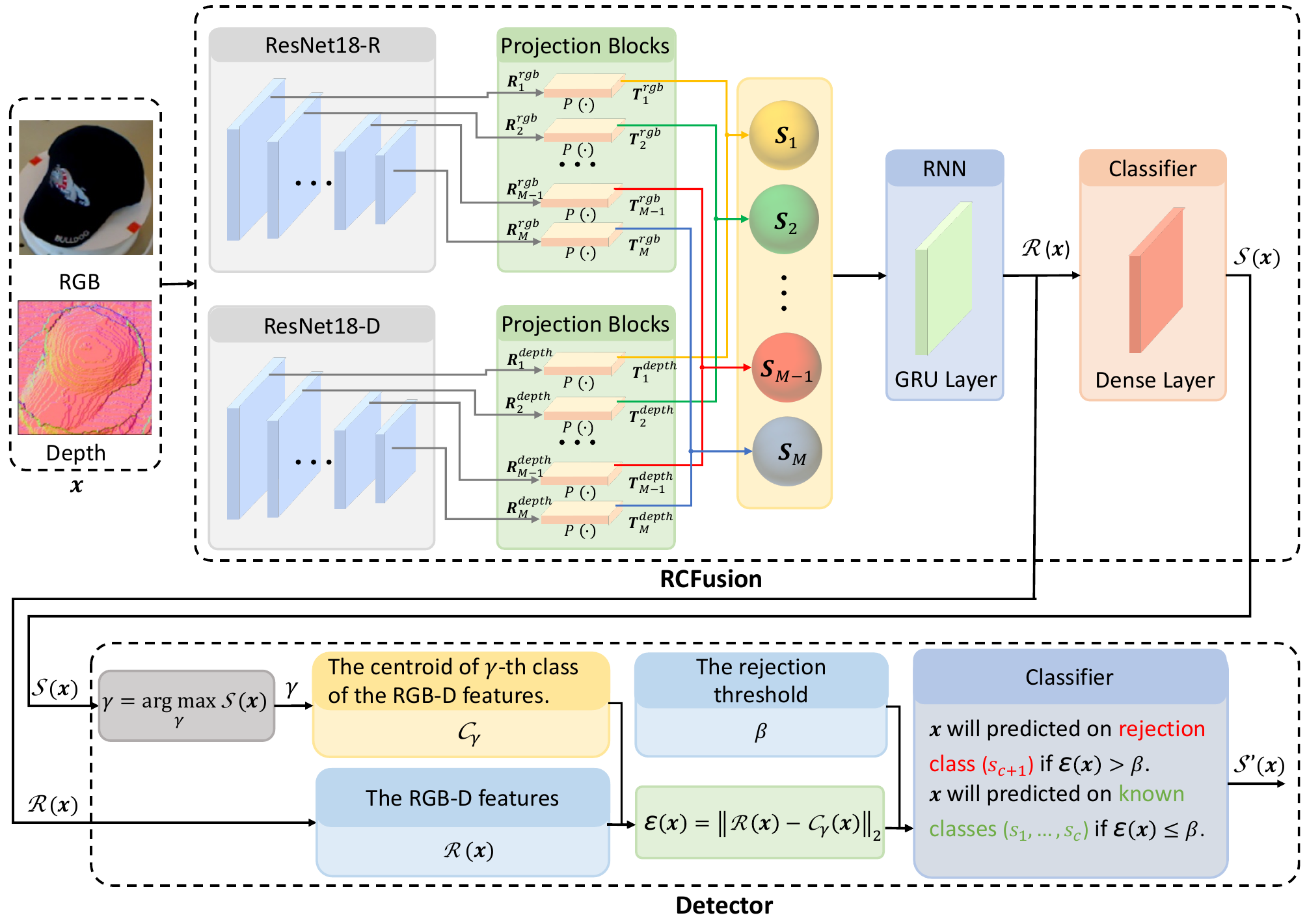}
    \caption{Adversarial examples detection framework architecture. It consists of RCFusion and the detector. RCFusion consists of two streams of CNN (\eg ResNet-18) employed to extract RGB and depth features at multiple levels of abstraction. The outputs of corresponding hidden layers are projected into a common space, concatenated, and sequentially fed to an RNN to obtain a compact RGB-D feature used by a classifier for the final classification. The detector will reject (accept) the input sample $\vct x$ if $\mathcal{E}(\vct x)$ is greater than (less than or equal to) the rejection threshold $\beta$.}
    \label{fig:rcfusion}
\end{figure*}

\section{Background}
\label{sec:background}
In the following section, we introduce the main concepts we will leverage in this work. 
We start by describing how RGB-D object recognition systems function (Section~\ref{subsec:rgbd}), and we discuss the threats posed by adversarial examples and adversarial patches (Section~\ref{subsec:adv}).
\subsection{RGB-D Object Recognition Systems}
\label{subsec:rgbd}
The idea of combining both colored RGB information and depth has been introduced in literature by Socher et al.~\citep{socher2012convolutional}, where the authors create a classifier of RGB-D images that employ a CNN and RNN to obtain the deep features, which are fed to an SVM to produce the final classification.
Instead, Eitel et al.~\citep{eitel2015multimodal} fuse RGB and depth features before computing the classification task by combining the information of the two streams in the pre-last layer of the neural network.
In this work, we consider the state-of-the-art architecture proposed in ~\citep{loghmani2019recurrent} for RGB-D object recognition and called recurrent convolutional fusion network (RCFusion). 
As shown in Fig.~\ref{fig:rcfusion}, RCFusion was designed by using two streams of convolutional networks (CNN), with the same architecture as ResNet-18~\citep{he2016deep} and pre-trained on the ImageNet, to extract RGB and depth features at different levels of the networks. 
While the RGB information does not require any particular pre-processing, depth information is not used "as-is" as a scalar number acquired by a sensor. It is post-processed to produce a colorized image.
Each pixel value of this image represents not a color but the normal of the surface acquired by the sensor.
The outputs of the corresponding hidden layers (the first of the network trained on the RGB images, with the first of the network trained on the depth image, the second ... ) are then projected into a common space, concatenated, and sequentially fed to a recurrent neural network (RNN) to obtain a compact RGB-D feature that is used by a classifier for the final classification. Let ResNet18-R and ResNet18-D represent the CNN for extracting RGB and depth features, and the output of the $i$-th layer of ResNet18-R and ResNet18-D are $ \vct R_i^{rgb}$ and $ \vct R_i^{depth}$, with $i = 1,2,\ldots, M$, and $M$ is the total number of layers of ResNet-18. Given that the dimension of the features obtained from different hidden layers of the same network is different, we apply the projection block $P(\cdot)$ proposed in ~\citep{loghmani2019recurrent} to transform a volumetric input into a vector of dimensions. Let the transformed RGB and depth features of $i$-th layer denote with $ \vct T_i^{rgb}$ and $ \vct T_i^{depth}$, \ie $\vct T_i^{rgb}=P(\vct R_i^{rgb})$ and $\vct T_i^{depth}=P(\vct R_i^{depth})$, and then concatenate the transformed RGB and depth features of $i$-th layer to form $\vct S_i = [\vct T_i^{rgb}; \vct T_i^{depth}]$. To create a compact multi-modal representation, we sequentially fed the set $\{\vct S_1, \vct S_2, \ldots ,\vct S_M\}$ to an RNN. 
There are two ways to instantiate RNN, as presented in~\citep{loghmani2019recurrent}, e.g., Gated Recurrent Unit (GRU), and Long-Short Term Memory (LSTM). 
Since the performance of GRUs and LSTMs are comparable, and GRUs have fewer parameters than LSTMs, applying a GRU layer to handle multimodal features is therefore convenient. Then a dense layer is combined with RNN to predict the final label.

\subsection{Adversarial Examples and Patches}
\label{subsec:adv}
While machine learning technologies are currently wide-spreading across many different domains, we are witnessing a rapid growth of studies proving their weaknesses against multiple and rapid-evolving threats at training~\citep{Cina2022Survey,Cin2022MachineLS} and at test time~\citep{BIGGIO2018317}.
Test time attacks formalize the presence of attackers that can compute \emph{adversarial examples} $\vct \delta^\star$, carefully-crafted perturbations applied to input samples designed to have them misclassified by the target model as the attacker desires~\citep{biggio2013evasion, szegedy2015intriguing}. For example, to have a malicious application misclassified as a legitimate one. 
Adversarial examples are the result of an optimization problem formulated as follows:
\begin{equation}
    \begin{aligned}
        \vct \delta^\star = \argmin_{\parallel \vct \delta \parallel_p \leq \epsilon} \mathcal{L}(\vct x + \vct \delta, y; \vct \theta)
    \end{aligned}
    \label{eq:adv_examples}
\end{equation}
where:
\begin{equation}
    \mathcal{L} = s_y(\vct x + \vct \delta) - \max_{j \notin \{y\}} s_j(\vct x + \vct \delta)
    \label{eq:loss}
\end{equation}
$\vct x$ is an input sample, $y \in \mathcal{Y} = \{1,\ldots,c\}$ is the true label of $\vct x$, $\vct \theta$ corresponds to the parameters of the target model, $\vct \delta$ is the adversarial perturbation, and $s_j(\vct x + \vct \delta)$, $j \in \{1,\ldots,c\}$ is the $j$-th output predictions score of the target model on the adversarial sample $\vct x + \vct \delta$. The constraint $\parallel \vct \delta \parallel_p \leq \epsilon$ is an $\ell_p$-norm constraint imposed to preserve stealthiness of the attack~\citep{Cina22Clustering}. Typical norms used for crafting adversarial examples are $\ell_1$, $\ell_2$, and $\ell_\infty$, for which efficient projection algorithms exist ~\citep{duchi08}.
However, since these manipulations are applied to all the pixels of an image, it is impossible to replicate them in a real-life scenario where a camera is looking at a scene.
To accomplish an attack in the described way, the attacker would need to either directly act on the camera sensor or tamper with the images before they are sent to the machine learning model.

A more realistic threat model to image classifiers is posed by adversarial patches: contiguous chunks of pixel values optimized to steer the decision toward a class decided by the attacker. These patches can be physically printed and placed on objects acquired by the camera~\citep{brown2017adversarial}.
The creation of adversarial patches amounts to solving an optimization problem similar to Eq.~\ref{eq:adv_examples}, described as follows:
\begin{equation}
    \vct \delta^\star = \argmin_{\parallel \vct \delta \parallel_p \leq \epsilon} \mathbb{E}_{\mat A \sim \mathcal{T}} \mathcal{L}(\vct x \oplus \mat A \vct \delta), y; \vct \theta),
    \label{eq:patches}
\end{equation}
where the adversarial patch $\vct \delta$ is applied to the input image $\vct x$ with random affine transformations $\mat A$ drawn from $\mathcal{T}$.
The operator `$\oplus$' is defined as: $\vct x \oplus \mat A\vct \delta = (\vct 1 - \vct \mu) \circ \vct x + \vct \mu \circ \mat A\vct \delta$, where the operator `$\circ$' means element-wise vector multiplication, $\vct \mu$ is a mask with the same size of the input data $\vct x$, and its components are ones where the patch should be applied and zeros elsewhere~\citep{karmon2018lavan}.
Eq.~\ref{eq:patches} can be minimized by the Algorithm~\ref{alg:pgd_patch} to create a perturbation that is still effective regardless of its position, rotation, and scale inside the image, hence mimicking acquisitions of the scene containing the image and the patch through a camera. 
In this paper, we leverage a simplified version of Eq.~\ref{eq:patches}, where we do not apply affine transformations to the patch but compute a single adversarial patch for each sample used at test time during the attack. This is the worst-case scenario for the defender, as each patch is optimized specifically for the image to which it is applied and is thus obviously more effective than a single patch computed to work on multiple images. 
Lastly, we only consider patch attacks that target the RGB channel of the input samples since modifying the depth information would require the attacker to possess the capability of assembling physical objects to mimic the adversarial perturbation, which would make them difficult to apply in real-world scenarios and costly to generate. The 3D printers that have high color precision (needed to allow the attacker to manipulate the colors) are, in fact, still quite expensive.

\begin{algorithm}[t]
	\caption{PGD-based Adversarial Patch Attack}
	\begin{algorithmic}[1]
		\REQUIRE $\vct x$: the input sample; $\mat A$ is a random affine transformations drawn from $\mathcal{T}$; $\eta$: the step size; $\Pi$: a projection operator on the $\ell_p$-norm constraint $\|\vct\delta\|_p \leq \epsilon$; $n$: the number of iterations.
		\ENSURE $\vct \delta$: the adversarial patch.
		\STATE $\vct \delta \gets \vct 0$
		\FOR{$i \in [0, n]$}
                \STATE $\vct \delta \leftarrow \vct \delta - \eta \nabla_{\vct \delta} \mathcal{L}(\vct x \oplus \mat A \vct \delta, y ;\vct \theta)$
                \STATE $\vct \delta \leftarrow \Pi(\vct \delta)$
		\ENDFOR
		\RETURN $\vct \delta$
	\end{algorithmic}
	\label{alg:pgd_patch}
\end{algorithm}

\section{The Robustness of RGB-D Object Recognition}
\label{sec:robustness}
The high vulnerability of RGB-D systems to adversarial perturbations due to the presence of RGB features had already been noted for object detectors in \citep{tu21-acrl}. 
However, as far as we know, no one has explained its underlying reasons. 
To bridge this gap, we analyze the internal structure of RGB-D models by computing the similarity between layers. 
We conjecture that models with high similarities between their hidden layers learn simpler decision functions; therefore, they tend to be more robust against adversarial manipulation of input data.
Conversely, when hidden layers are dissimilar, the underlying decision function is more complicated, creating holes in the decision space where adversarial examples lie.
In the following, we revisit the Centered Kernel Alignment (CKA)~\citep{kornblith2019similarity} as the similarity measure we use in this paper to determine the similarity between the hidden layers of neural networks. 
We then exploit the CKA similarity matrices to explain why these systems are more vulnerable to perturbation on the RGB features of the input rather than on the depth descriptor.\smallskip

\myparagraph{HSIC.} Before delving into the details of CKA, we introduce the Hilbert-Schmidt Independence Criterion (HSIC), subsequently used for computing the CKA measure.
Introduced by Gretton et al.~\citep{gretton2005measuring}, HSIC is a useful method for testing if two random variables are independent. 
Formally, suppose $\vct X \in \mathbb{R}^{m \times p_1}$ and $\vct Z \in \mathbb{R}^{m \times p_2}$ are the output features of the two hidden layers, having respectively $p_1$ and $p_2$ neurons, for $m$ input samples. We then denote with $\vct x_i$, $\vct x_j$ ($\vct z_i$, $\vct z_j$) the $i$-th and $j$-th entries in matrix $\vct X$ ($\vct Z$), respectively representing the features representation for the $i$-th and $j$-th samples.
We finally define with $\vct K_X=\{K_X^{ij}\}_{i,j}$ and $\vct K_Z=\{K_Z^{ij}\}_{i,j}$ the symmetric kernel matrices used to evaluate the similarities of features abstracted from the two layers with $p_1$ and $p_2$ neurons separately. For computing the two matrices, we used two distinct kernels: a linear kernel function, where $ K_X^{ij} = {\vct x_i}{\vct x_j}^{\rm T}$, $ K_Z^{ij} = {\vct z_i}{\vct z_j}^{\rm T}$; and the Radial Basis Function (RBF), where $K_X^{ij} = {\rm exp}(-\frac{{\Vert \vct x_i - \vct x_j \Vert }^2}{2\sigma^2})$, $K_Z^{ij} = {\rm exp}(-\frac{{\Vert \vct z_i - \vct z_j \Vert }^2}{2\sigma^2})$ and $\sigma$ is chosen as a fraction of the median distance between features. 
Obviously, these two kernel functions satisfy $K^{ij}=K^{ji}$. Based on the empirical estimator of HSIC(\citep{gretton2005measuring}, Definition 2), we can obtain Eq.~\eqref{eq:hsic}:

\begin{equation}
\label{eq:hsic}
    {\rm HSIC}(\vct X, \vct Z) = \frac{1}{(n-1)^2} \mat{tr}(\vct K_X \vct H \vct K_Z \vct H),
\end{equation}
where centring matrix $\vct H = \vct I_n - \frac{1}{n} \vct 1 \vct 1^{\rm T}$, $\vct I_n$ is the identity matrix of size $n \times n$, and $\vct 1$ is $n \times 1$ vector of all ones. \smallskip

\myparagraph{CKA.} HSIC is invariant to orthogonal transformations of the representations and, by extension, to permutation of neurons, but it is not invariant to scaling of the original representations. CKA~\citep{kornblith2019similarity} further normalizes HSIC to produce a similarity index between 0 and 1 that is invariant to isotropic scaling. Formally, the CKA similarity between two matrices $\vct X$ and $\vct Z$ is defined as: 

\begin{equation}
\label{eq:cka}
    {\rm CKA}(\vct X, \vct Z) = \frac{{\rm HSIC}(\vct X, \vct Z)}{\sqrt{{\rm HSIC}(\vct X, \vct X){\rm HSIC}(\vct Z, \vct Z)}}.
\end{equation}

For the Eq.~\eqref{eq:cka}, it is not difficult to see that ${\rm CKA}(\vct X, \vct Z)=1$ when $\vct X= \vct Z$, namely, $\vct X $ and $\vct Z $ are the feature representation from the same layer. Besides, ${\rm CKA}(\vct X, \vct Z)=0$ when ${\rm HSIC}(\vct X, \vct Z) = 0$, this means that $\vct X $ and $\vct Z $ are independent of each other.

\section{Defending RGB-D Classifiers}
\label{sec:defense}
In the above section, we presented our methodology for inspecting which are the vulnerable component defining an RGB-D object recognition system. 
We here present our defensive mechanism and how we assess its robustness against adaptive attacks, i.e., attacks specifically designed to target a given defense. 
In the following, we denote with $\vct x$ and $\tilde{\vct x}$ two generic samples taken respectively from the test and training set. We then denote the output predictions score of RCFusion, trained on the RGB and depth features, with $\mathcal{S}(\vct x) = [s_1(\vct x), \ldots,s_c(\vct x)] \in \mathbb{R}^{1 \times c}$, where $s_i(\vct x) \geq 0$ and $\sum_{i=1}^{c} s_i(\vct x) = 1$. A complete summary of the notation and symbols used throughout the paper is reported in Table~\ref{tab:glossary} (see~\ref{appendix:notation}).\\

\subsection{Reject-based Detection}
\begin{figure}[t]
	\centering
    \includegraphics[trim=0 0 0 0, clip, width=.40\textwidth]{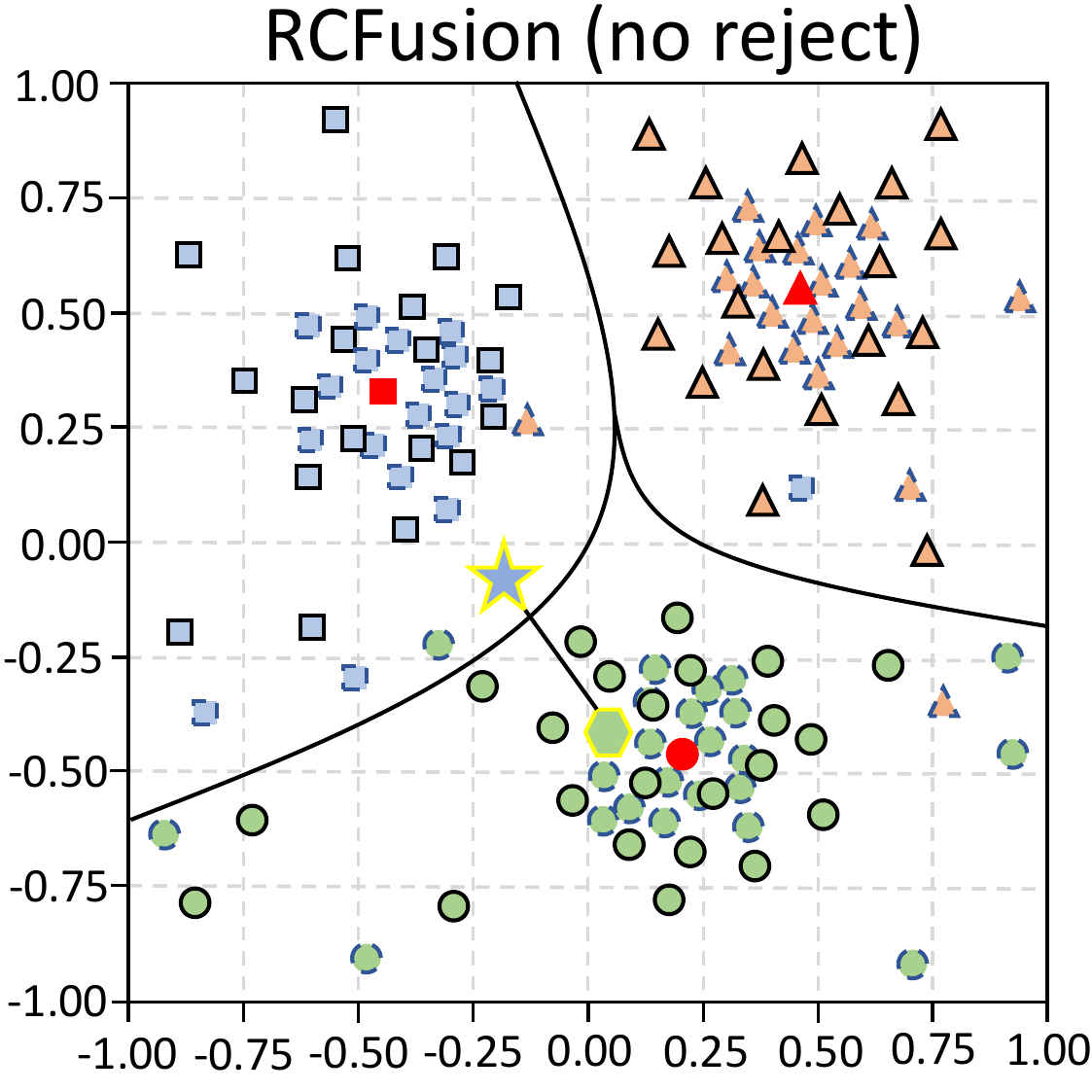}
	\includegraphics[trim=0 0 0 0, clip, width=.40\textwidth]{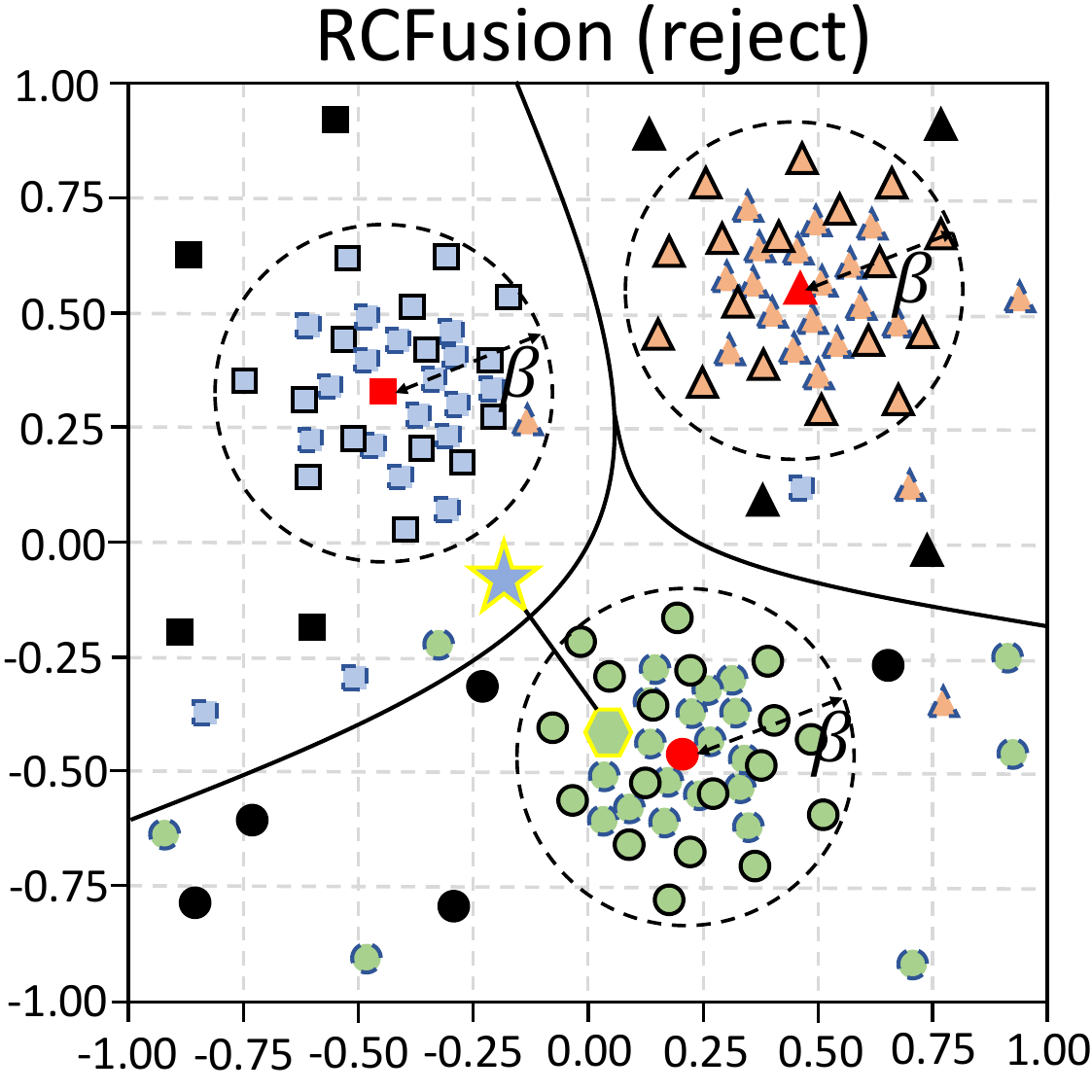}
	\caption{
	Visual representation of evasion attack on 3-class bi-dimensional classification problem: left without defense, right with reject-based defense. Blue dotted lines for final depth features, black solid lines for final RGB-D features, red dots for centroids, and black dots for rejected samples. The rejection threshold is shown as a black dotted circle. Green hexagon for the initial sample and a blue star for the adversarial sample. Defense correctly rejects the adversarial sample, while without defense, it was wrongly classified as belonging to the blue class.}
 \label{fig:attack}
\end{figure}
\aedit{The only defense~\citep{wang2022adversarial} that has been proposed to secure RGBD-base systems is based on adversarial training~\citep{madry2018towards}. Defenses based on adversarial training present two problems: (i) they increase the margin between the classes; however, if the perturbation of the attacker can inject is slightly higher than the margin, they are ineffective; (ii) they require generating many adversarial examples during training, which is computationally demanding as the generation of each adversarial example requires multiple forward and backward pass (and this should be done for all the samples of the training set or at least the subset considered). 
Therefore, we propose a defense based on a detector. The underlying idea of our defensive method is to estimate the distribution of unperturbed training points at different network layers and reject anomalous samples that may be incurred at test time, including adversarial examples. Specifically, our defensive mechanism rejects out-of-distribution samples at test time by looking at their RGB and depth information, and as far as a sample moves away from class centroids, classifier support decreases to zero. Thus, this defense can detect adversarial examples that are highly perturbed and do not require generating adversarial examples at training time. The operations required at training time are: (i) computing, for each class, its centroid in RGB-D space; (ii) finding the rejection threshold, which requires computing some distances in RGB-D space. Both these operations can be performed by computing just once (and thus with a single forward pass) the RGB-D features of the training samples (or the subset considered). Therefore, the proposed approach is rather more efficient at training time than adversarial training. Whereas at test time, both do not require expensive operations. Adversarial training does not require any other operation than the standard classification. At the same time, our defense requires computing the distance in RGB-D space between the considered samples and the centroid of the predicted class (of which the RGB-D features have already been computed and stored at training time).
The architecture of our defense mechanism is depicted in Fig.~\ref{fig:rcfusion}, which assumes the defender has an already-trained classifier to be protected against adversarial examples.}

For our defense mechanism to work, we compute the centroid of the final RGB-D features for each class, as shown in Fig.~\ref{fig:attack}, and then reject anomalous samples whose RGB-D representation is far from the centroid. Without loss of generality, our approach uses the $\ell_2$ distance between the RGB-D features of the class-centroid and the RGB-D features of the input sample. 
Formally, we denote the final RGB-D features of RCFusion trained on the RGB and depth parts with $\mathcal{R}(\vct x) = [r_1(\vct x), \ldots, r_{a}(\vct x)] \in \mathbb{R}^{1 \times a}$, being $a=100$ the dimensionality of the output features of the RNN layer. For each class $\gamma$ in the training set, we then compute its corresponding centroid $\mathcal{C}_\gamma$ with respect to their RBG-D feature as:
\begin{equation}
    \mathcal{C}_\gamma = \frac{1}{n_{\gamma}}\sum\limits_{k=1}^{n_{\gamma}} \mathcal{R}(\tilde{\vct x}_k^\gamma)
\end{equation}
where $n_{\gamma}$ is the number of samples that belonging to class-$\gamma$, $\tilde{\vct x}_k^\gamma$ is the $k$-th sample which from the class-$\gamma$ of the training set.

We then define \textit{anomaly score} $\mathcal{E}$ for the test sample $\vct x$ as:
\begin{equation}
\mathcal{E}(\vct x)=\parallel \mathcal{R}(\vct x) -\mathcal{C}_\gamma(\vct x) \parallel_2 
\end{equation}
being $\gamma=\argmax_{\gamma} \mathcal{S}(\vct x) \in [1, c]$ the predicted label of RCFusion trained on the RGB and depth parts.

Finally, the detector 
will reject samples if $\mathcal{E}(\vct x)$ is greater than the rejection threshold $\beta$, whose optimal value can be found with the Algorithm~\ref{alg:threshold}. According to this rule, we define the output predictions scores of RCFusion with the detector as:  $\mathcal{S}'(\vct x) = [(1-s_{c+1})s_1(\vct x), \ldots,(1-s_{c+1})s_c(\vct x), s_{c+1}(\vct x)] \in \mathbb{R}^{1 \times {(c+1)}}$ where the rejection class $c+1$ is defined as follows:
\begin{equation}
\label{eq:sss}
    s_{c+1}(\vct x) = 
    {\begin{cases}
    1, \qquad& \ if \ \mathcal{E}(\vct x) > \beta\\
    0, \qquad& \ if \ \mathcal{E}(\vct x) \leq \beta 
    \end{cases}}.
\end{equation}
The test samples are then assigned to the class for which the value of $\mathcal{S}'(\vct x)$ is higher. 
A test sample $\vct x$ is thus assigned to the rejection class $c+1$ when $\mathcal{E}(\vct x) > \beta$; otherwise, it is assigned to the class with the highest likelihood in the softmax output.

\subsection{Attacking the Defended System}

\begin{algorithm}[t]
	\caption{Compute the Rejection Threshold}
	\begin{algorithmic}[1]
		\REQUIRE $T$: the number of iterations; $\rho$: the step size; $n$: the number of rejected samples; $N$: the total number of training samples; $\tilde{\vct x}_k$: the $k$-th training sample; $\mathcal{R}(\tilde{\vct x}_k)$: the final RGB-D features of RCFusion trained on the RGB and depth parts; $\mathcal{C}_\gamma(\tilde{\vct x}_k)$: the centroid of $\gamma$-th class of the final RGB-D features; $\gamma$: the predicted label of RCFusion trained on the RGB and depth parts; $r$: the False Positive Rate (FPR).
		\ENSURE $\beta$: the rejection threshold.
		\FOR{$i = 1 : T$}
		\STATE $\beta_i = \rho \times i, n=0$
		\FOR{$k=1:N$}
		\IF{$\parallel \mathcal{R}(\tilde{\vct x}_k) -\mathcal{C}_\gamma(\tilde{\vct x}_k) \parallel_2 > \beta_i$}
		\STATE $n \gets n + 1$
		\ENDIF
		\ENDFOR
		\IF{$ \frac{n}{N} > r$}
		\STATE $\text{continue}$
		\ELSE
		\STATE $\beta \gets \beta_i$
		\STATE $\text{break}$
		\ENDIF
		\ENDFOR
		\RETURN $\beta$
	\end{algorithmic}
	\label{alg:threshold}
\end{algorithm}

When a defense is based on a detector to reject the adversarial examples, a defense-unaware attack may craft adversarial examples belonging to rejection regions, making it very difficult to evade such defense (Fig.~\ref{fig:attack})~\citep{sotgiu2020deep}. To perform a fair robustness evaluation of the proposed defense method, an adaptive defense-aware attack is required. Therefore, we formulate an adaptive white-box attack suitable for assessing the adversarial robustness of the proposed rejection-based. 

Given a sample $\vct x$, the attacker can optimize a maximum-allowed $\epsilon$-sized adversarial perturbation obtaining the defense-aware adversarial perturbation $\vct \delta^\star$, by solving the following constrained optimization problem:

\begin{equation}
    \vct \delta^\star = \argmin_{\parallel \vct \delta \parallel_p \leq \epsilon} \mathcal{L}_d(\vct x + \vct \delta, y; \vct \theta)
    \label{eq:defense_aware_examples}
\end{equation}

where $\parallel \vct \delta \parallel_p \leq \epsilon$ is an $\ell_p$-norm constraint. The formulation of the adaptive attacks is similar to the one seen in Eq.~\eqref{eq:adv_examples}, with the only difference that now the target loss $\mathcal{L}_d$ takes into consideration also the detector defense the attacker aims to evade. Formally, we define $\mathcal{L}_d$ as follows:
\begin{equation}
    \begin{aligned}
    \mathcal{L}_d &
    = s_y(\vct x + \vct \delta) - \max_{j \notin \{y, c+1\}} s_j(\vct x + \vct \delta) 
    \end{aligned}
\end{equation}
where $c+1$ is the rejection class. Compared to Eq.~\eqref{eq:loss}, the attacker enforces not only that the class predicted for the adversarial example does not match the true label but also that it does not match the rejection class.

In the context of image classification, the solution of the minimization problem mentioned before produces a perturbation that, applied to the pixel values of the input image, forces the target model to predict the sample to a class that is different from the true class.
To achieve this error-generic (untargeted) evasion, the attacker should minimize the output of the true class and maximize the output of one competing class (excluding the reject class). It is worth noting that this algorithm performs a strong maximum-confidence evasion attack (rather than searching for a minimum-distance adversarial example).
While in this work, we focus only on \emph{untargeted} attacks, the proposed formulation can also be easily extended to account for error-specific (targeted) evasion. Note that the \emph{targeted} attacks requires the model to misclassify the sample to the class decided a priori by the attacker, which can be written similarly to Eq.~\eqref{eq:defense_aware_examples} by using the target label $y_t$ instead of $y$, and inverting the sign of the loss function~\citep{melis17-vipar}.

Moreover, to solve the optimization problem above, given that the Eq.~\eqref{eq:sss} is a step-function and non-differentiable, we apply:
\begin{equation}
    s_{c+1}(\vct x)=\frac{1}{1+\exp{(-\lambda(\mathcal{E}(\vct x)-\beta))}}
    \label{eq:approximation}
\end{equation}
to implement the loss function exploited by the attacker to compute the adversarial examples in our experiment.

\section{Experimental Analysis}
In our experimental analysis, we consider two different RGB-D datasets to perform multi-modal computer vision classification tasks. Our analysis has three fundamental objectives: (i) investigating the robustness of RCFusion to detect which are the most vulnerable features; (ii) interpreting the previous results by inspecting the similarity between hidden layers of RCFusion; and (iii) testing the robustness of the proposed defense against defense-unaware and adaptive attackers. 
In the following, we define the experimental setup adopted in our empirical analysis to foster the reproducibility of our results, and we then present and analyze our findings.
\subsection{Experimental Setup}
\myparagraph{Datasets.} 
We conduct our experiments by choosing two datasets, i.e., RGB-D Object Dataset~\citep{Lai11ORGBDObjectDataset} and OCID~\citep{Suchi19OCID}, where data dimensionality and the number of classes are different, thus making our setup more heterogeneous and challenging.\smallskip

\noindent \textit{RGB-D Object Dataset}~\citep{Lai11ORGBDObjectDataset}\footnote{\url{http://rgbd-dataset.cs.washington.edu/}} contains 300 common household objects taken from multiple views organized into 51 categories with a total of 207,920 RGB-D images. 
It was sampled using a Kinect-style 3D camera that records synchronized and aligned $640 \times 480$ RGB and depth images at $30$Hz. 
Due to the massive dataset size, we subsampled it by extracting only every fifth frame, thus obtaining $41,877$ RGB-D images.
We run our experiment based on ten cross-validation splits: one object instance per class is used for testing, and training is performed on the remaining 249 $(300-51)$ instances, where each split consists of roughly 35,000 training images and 7,000 images for testing.\smallskip

\noindent \textit{Object Clutter Indoor Dataset (OCID)}~\citep{Suchi19OCID}\footnote{\url{https://www.acin.tuwien.ac.at/en/vision-for-robotics/software-tools/object-clutter-indoor-dataset/}} comprises 96 fully built up cluttered scenes representing common objects organized in three subsets: ARID20, ARID10, and YCB10. The ARID20 and ARID10 subsets include cluttered scenes with up to 20 and 10 objects from Autonomous Robot Indoor Dataset (ARID) objects, respectively, whereas the ARID20 (ARID10) subset includes cluttered scenes with $3,180$ ($2,499$) RGB-D images. 
Moreover, the YCB10 subset includes cluttered scenes with up to 10 objects from YCB objects. The data capture diverse settings of objects, backgrounds, context, sensor-to-scene distance, viewpoint angle, and lighting conditions. In our experiment, we have chosen ARID20 (ARID10) as the training (testing) set. \smallskip

\begin{figure}[t]
    \centering
    \includegraphics[width=0.6\textwidth]{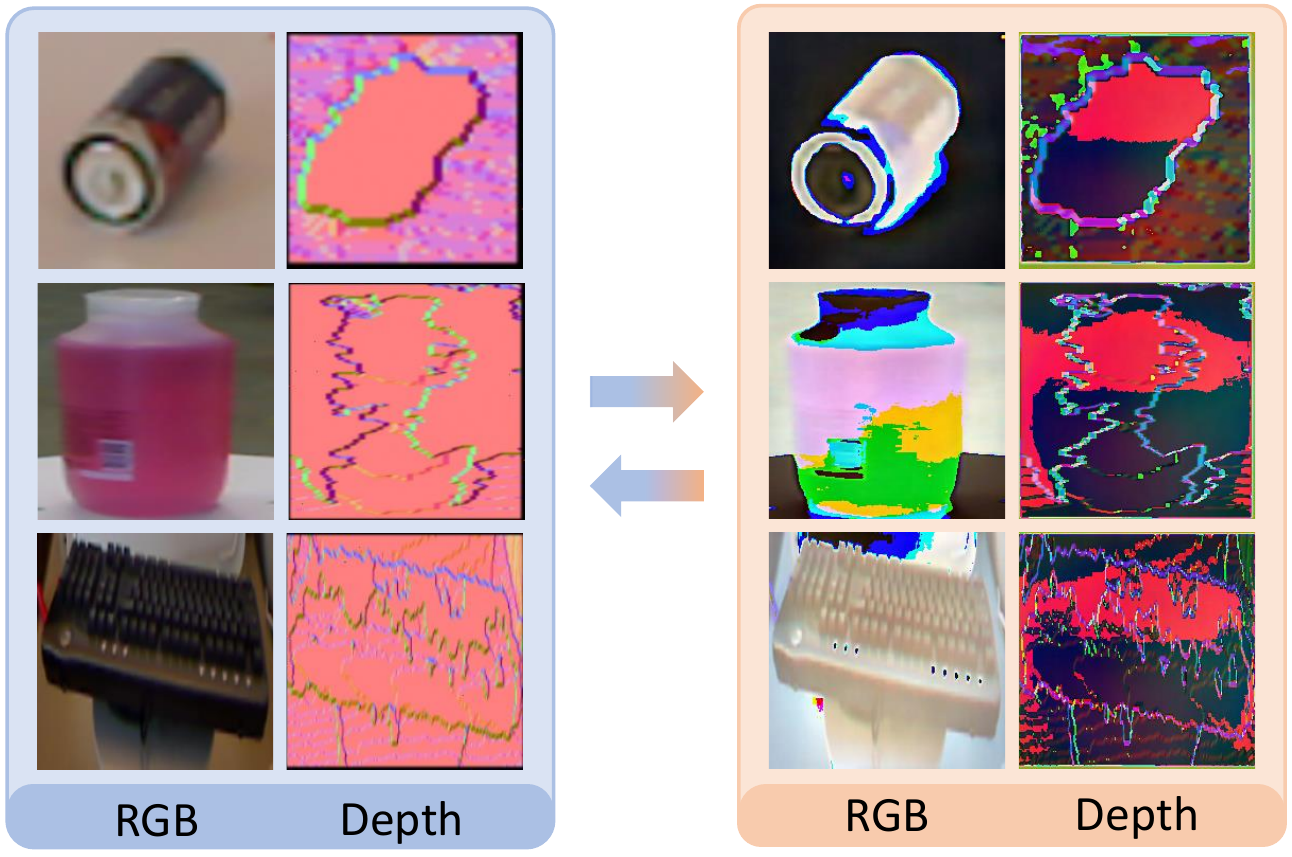}
    \caption{Samples in the preprocessing and inverse-preprocessing spaces, where the inverse-preprocessing (preprocessing) space is highlighted in blue (orange).}
    \label{fig:preprocessing}
\end{figure}

\myparagraph{Preprocessing.} \tocheck{To obtain the colorized depth images, we first normalize the original depth before proceeding to the colorization by adopting the method based on surface normal~\citep{Aakerberg2017DepthVP,bucci2018multimodal}. The resulting representation focuses on capturing structural information (e.g., object shapes, surface properties, and relative orientations) while being invariant to the distance to the camera or the total depth range~\citep{Aakerberg2017DepthVP}. }

For the preprocessing procedure, we convert RGB and colorized depth images from the RGB to the BGR space.
We then resize the images of BGR space as $256 \times 256 $ and subtract the mean values \footnote{\url{https://data.acin.tuwien.ac.at/index.php/s/RueHQUbs2JtoHeJ}} provided by Mohammad et al.~\citep{bucci2018multimodal}, and then apply a resize again to shrink images to match the input size of the considered model, which is $224 \times 224 $. 
For the inverse-preprocessing, we resize preprocessed images as $256 \times 256 $, and then add the mean values to the resized images, convert the images of BGR space to RGB space, and resize the images of RGB space to $224 \times 224 $. 
Finally, the results of preprocessed and inverse-preprocessed are shown in Fig.~\ref{fig:preprocessing}.

\myparagraph{Classifiers.} We train the model using RMSprop optimizer with batch size $64$, learning rate $0.0001$, momentum $0.9$, weight decay $0.0002$, projection depth $256$, and the number of memory neurons $a=100$~\citep{loghmani2019recurrent}.
We report in Table~\ref{tab:training_results} the performances of the trained models on different feature sets (RGB, depth, and RGB-D) on the RGB-D Object Dataset and OCID.
Also, we include the performance of another deep neural architecture, which is AlexNet~\citep{krizhevsky2012imagenet}, a convolutional neural network originally trained on ImageNet~\citep{deng2009imagenet}.
We use Alexnet as another backbone neural network for RCFusion alongside ResNet-18.
We also use the pre-trained ResNet-18 to classify the RGB images of the considered datasets and train the model using the RMSprop optimizer, where the batch size, learning rate, momentum, and weight decay are the one proposed by the original training of RCFusion`\citep{loghmani2019recurrent}.
We also report the accuracy of ResNet-18 in Table~\ref{tab:training_results}.\smallskip

\begin{table}[t]
\centering
\caption{Accuracy of RCFusion on ResNet-18 and AlexNet for object recognition on RGB-D Object Dataset and OCID.}
\label{tab:training_results}
\begin{tabular}{cccc}
\hline
Model & Input Part & \begin{tabular}[c]{@{}c@{}}RGB-D \\ Object Dataset\end{tabular} & OCID \\ \hline
\multirow{3}{*}{RCFusion on ResNet-18} & RGB-D & 95.04\% & 91.51\% \\
                         & RGB   & 88.70\%  & 87.02\% \\
                         & Depth & 82.03\%  & 40.35\% \\ \hline
\multirow{3}{*}{RCFusion on AlexNet} & RGB-D & 83.72\% & 65.22\% \\
                         & RGB   & 69.51\%  & 59.65\% \\
                         & Depth & 61.48\%  & 20.45\% \\ \hline
\multirow{2}{*}{ResNet-18} & RGB   & 87.65\% & 88.08\% \\
                           & Depth & 80.14\% & 35.05\% \\ \hline
\end{tabular}
\end{table}

\begin{figure*}[t]
    \centering
    \includegraphics[width=1\textwidth]{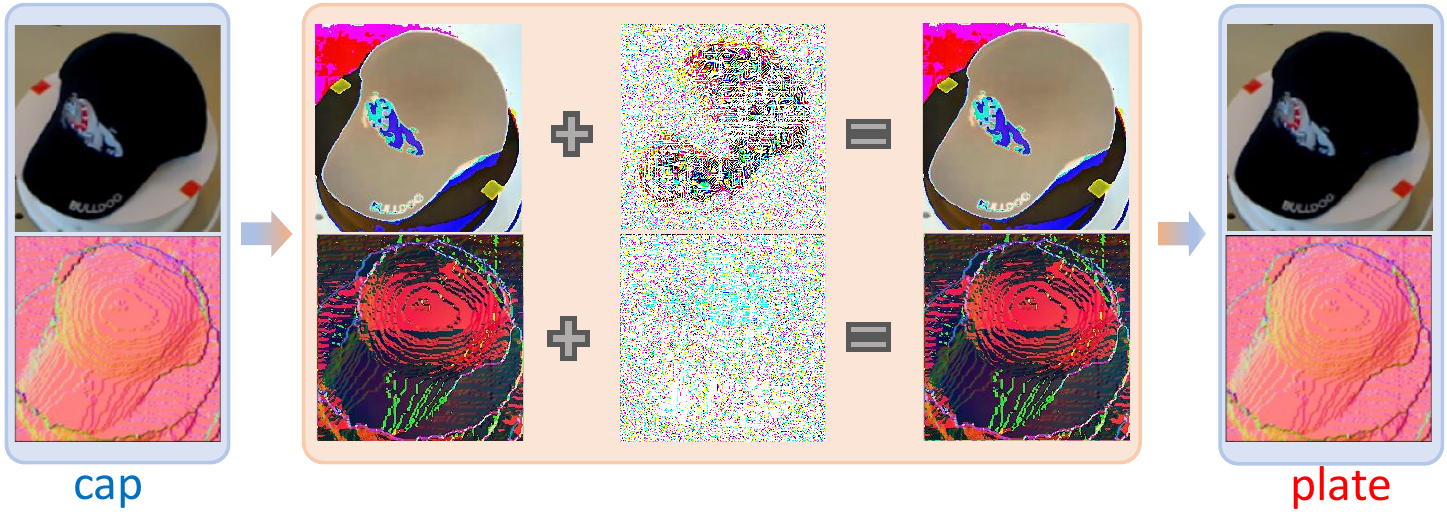}
    \caption{The scheme we use for modifying the full image to attack the RGB-D part of RCFusion, where the maximum perturbation is 0.3, and the inverse-preprocessing and preprocessing spaces are highlighted in blue and orange.}
    \label{fig:adversarial_attack_shown}
\end{figure*}
\begin{figure*}[t]
    \centering
    \includegraphics[width=0.99\textwidth]{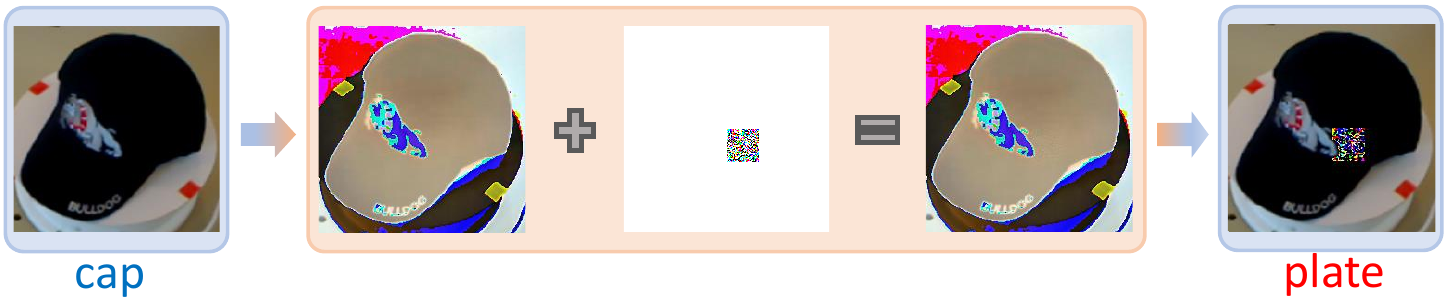}
    \caption{The scheme we use for modifying only a portion image to attack the RGB part of RCFusion, where the patch size is $35 \times 35$, and the inverse-preprocessing and preprocessing spaces are highlighted in blue and orange, separately.}
    \label{fig:adversarial_patch_shown}
\end{figure*}

\myparagraph{Adversarial Attack.} 
To evaluate the adversarial robustness, we test the trained networks with attacks that either jointly or separately target the RGB and depth parts.
We leverage the Adversarial Robustness Toolbox, (ART)\footnote{\url{https://github.com/Trusted-AI/adversarial-robustness-toolbox/}} from which we select the \emph{untargeted} $\ell_\infty$-norm version of Projected Gradient Descent (PGD) and PGD-based maximum-confidence patch attack.
PGD~\citep{madry2018towards} is first used to test the robustness when all the RGB and depth features are perturbed, as shown in Fig.~\ref{fig:adversarial_attack_shown}. 
Within this configuration, we perform 100 iterations, with a step size is $0.05$ for the RGB-D Object Dataset and OCID dataset.
Furthermore, to mimic a real-world scenario, we leverage the adversarial patch to attack against the RGB part, where the maximum perturbation $\epsilon=20$, the step size is $1$, and we let vary the patch size in $[0, 112]$. We provide an example of this attack schema in Fig.~\ref{fig:adversarial_patch_shown}.

\myparagraph{Parameter Setting.} \yzzedit{
For training RCFusion, we refer to the parameter setting provided by the author of ~\citep{loghmani2019recurrent} and, to obtain a fair comparison, we adjust them to obtain the performance of RCFusion on RGB-D Object Dataset and OCID as consistent as possible with the performance presented in ~\citep{loghmani2019recurrent}.} \aedit{We have fixed the only hyperparameter of our defense, namely the rejection threshold $\beta$, using Algorithm~\ref{alg:threshold}. This algorithm finds the rejection threshold appropriate to obtain the desired FPR ($r$), which in our experiments, we required to be equal to $10\%$  on the clean (unperturbed) samples. In the following, we discuss how the parameters of this algorithm should be set to obtain an appropriate threshold. The number of iterations ($T$) should be set large enough to allow it to converge. We have set it equal to $1e10$. Whereas the step size $\rho$ should be small because a small variation in the threshold can greatly impact the correspondent FPR. Thus we set it equal to $1e-5$.} 
\smallskip

\myparagraph{Performance Metrics.} We denote the original (undefended) performance of ResNet-18/AlexNet with ``ResNet-18''/``AlexNet'',  the original (undefended) performance of RCFusion with ``RCFusion'', \yzzedit{the adversarial training performance of RCFusion with ``RCFusion\_AT'',} the percentage of samples rejected by the detector with ``Rejection'' and the classifier defended with the proposed detector with ``Defended''.\smallskip

\myparagraph{Security Evaluation.} We compare the object detector and its robust model considering their security evaluation curves~\citep{BIGGIO2018317,Cin2021TheHA}, reporting classification accuracy against an increasing $\ell_\infty$-norm perturbation size $\epsilon$, used to perturb all the test samples. To set the scaling parameter $\lambda$ in Eq.~(\ref{eq:approximation}), we have tried different values, and we have found that the most appropriate to be 30.
There are two cases to calculate the accuracy of the robust detector: i) without attack (\ie, for $\epsilon=0$), the accuracy of the robust detector is computed as usual, but considering the errors of clean samples which are classified correctly but rejected; ii) under attack (\ie, for $\epsilon>0$), all the tested samples become adversarial examples, and we consider them correctly classified by the robust detector if they are classified either to the rejection class or their true class. We also report the rejection rates computed by dividing the number of rejected samples by the total number of tested samples. It is worth noting that the difference between the accuracy of the robust detector and rejection rate at each $\epsilon>0$ corresponds to the fraction of adversarial examples which are not rejected but still correctly assigned to their true class.\smallskip

\subsection{Experimental Results}
\label{sec:experiments_results}
We now discuss our empirical findings by reporting the robustness of RCFusion against adversarial attacks, analyzing the factors influencing its vulnerability, and the robustness of the proposed rejection-based defense.\\ 

\begin{figure}[t]
    \centering
    \includegraphics[trim=0 0 0 0, clip, width=.45\textwidth]{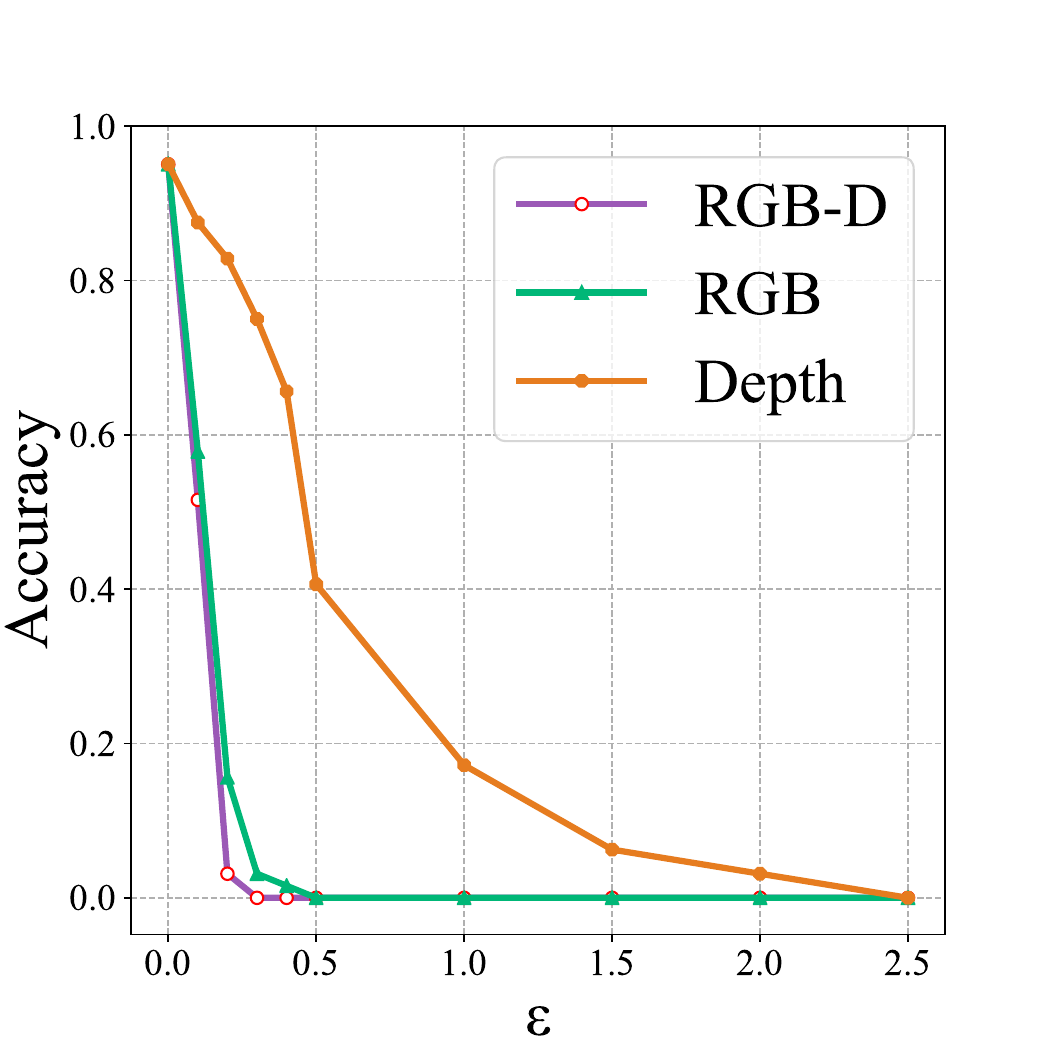}
	\includegraphics[trim=0 0 0 0, clip, width=.45\textwidth]{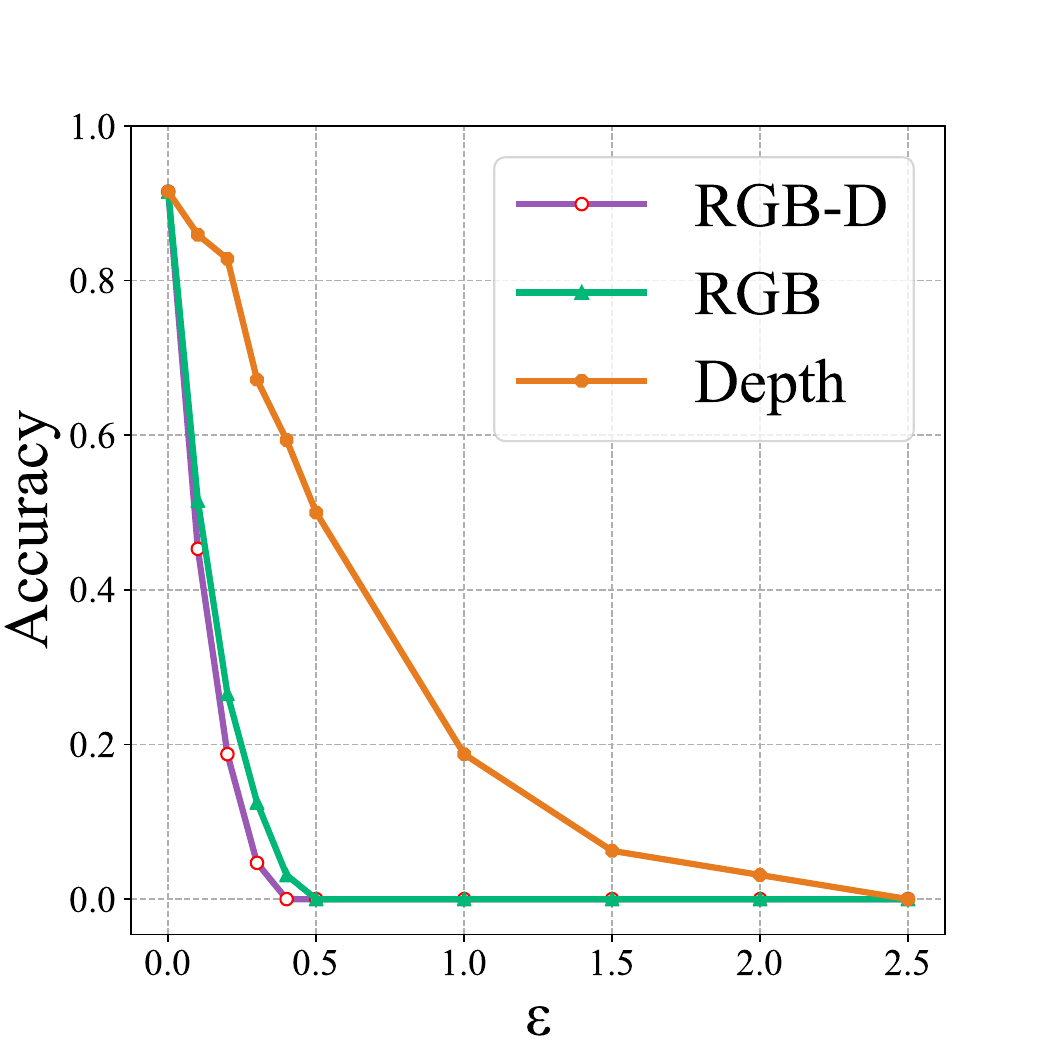}
    \caption{The robustness evaluation curve is computed by modifying the full image to attack the RGB, depth, and RGB-D parts of RCFusion for RGB-D Object Dataset (left) and OCID (right), respectively.}
    \label{fig:adversarial_attack}
\end{figure}
\begin{figure}[h!]
\centering
    \includegraphics[trim=0 0 0 0, clip, width=.45\textwidth]{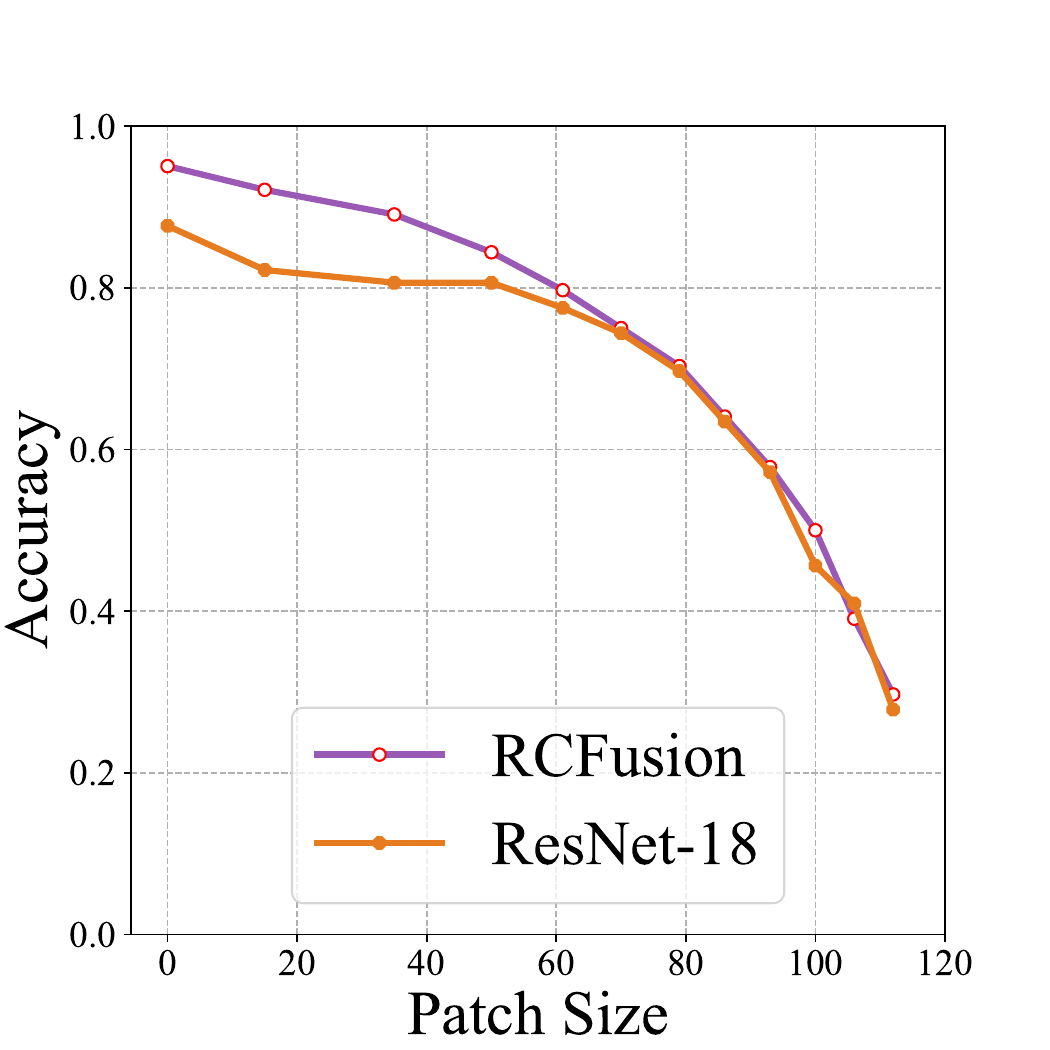}
	\includegraphics[trim=0 0 0 0, clip, width=.45\textwidth]{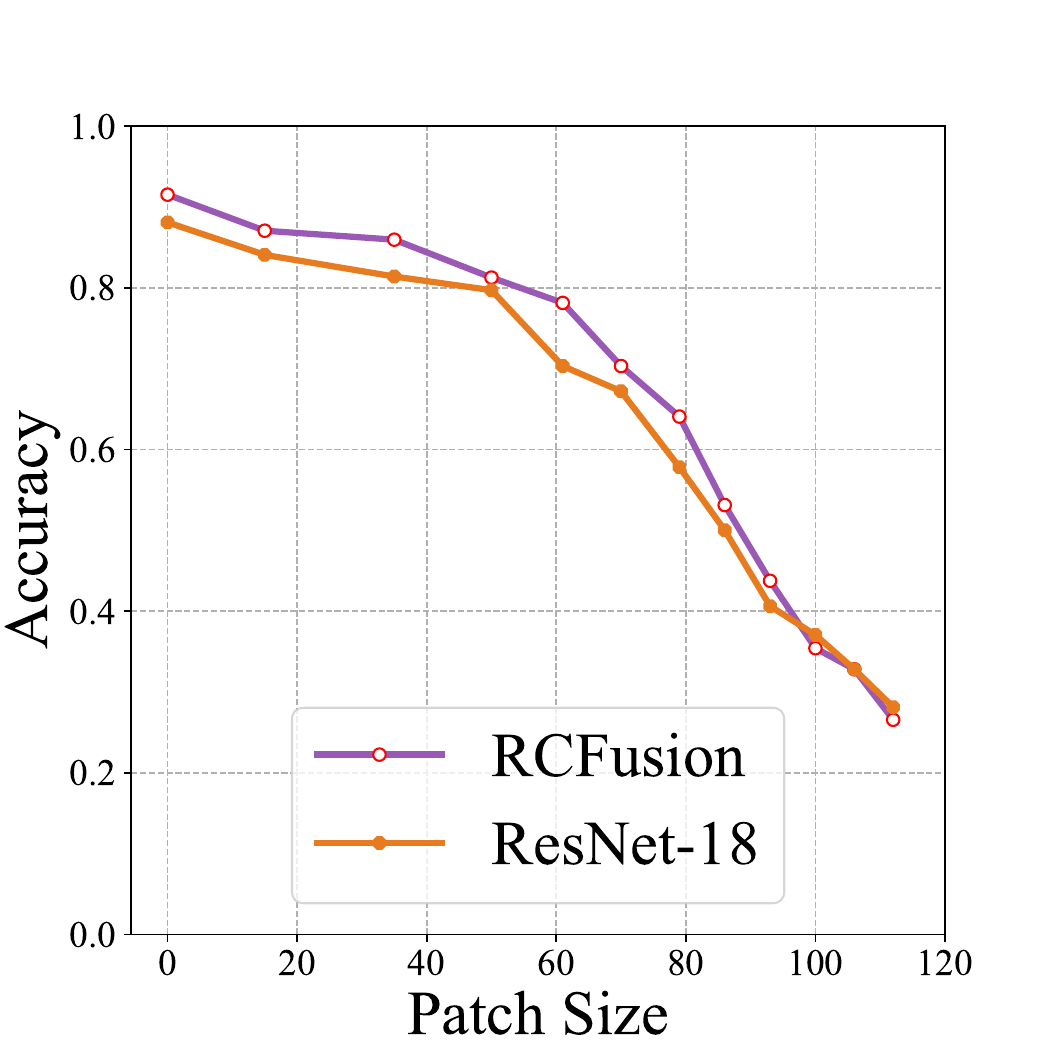}
    \caption{The robustness evaluation curve is computed by modifying only a portion image to attack the RGB part of RCFusion and ResNet-18 for RGB-D Object Dataset (left) and OCID (right), respectively.}
    \label{fig:adversarial_patch}
\end{figure}
\myparagraph{Robustness of RCFusion.} We here empirically investigate the robustness of RCFusion on the maximum perturbation $\epsilon$ following the attack pipeline depicted in Fig.~\ref{fig:adversarial_attack_shown}  To this end, we apply PGD~\citep{madry2018towards} separately against the RGB and depth feature only, and then we apply PGD against the RGB and depth feature combined. We report their results against RCFusion in Fig.~\ref{fig:adversarial_attack} when considering the RGB-D Object Dataset and OCID datasets.

The attack against both RGB and depth parts shows that the model is not robust against adversarial manipulations, as its accuracy drops to zero with a small perturbation budget.
Interestingly, this result is also achieved by computing PGD against the sole RGB part.
On the contrary, when PGD is applied against the depth part only, the attack needs a higher perturbation budget to drop the accuracy to zero.
From these results, it emerges that the \textit{performance of the whole RCFusion network is being hurt more by the RGB information than the depth information}. \aedit{We have carried out some experiments to investigate why, and we will present them later on in this section.}
However, the depth information alone, even if it can be expressive enough to exhibit more robustness than RGB information, usually leads to poor accuracy (as previously shown in Table~\ref{tab:training_results}). Therefore, training the classifier only on the depth features could not be considered a solution to obtain a classifier that is more robust to perturbations of the input. The big difference in the accuracy is given by the fact that the OCID dataset contains many objects that are distinct but have almost the same shape, like the ones belonging to the classes ``ball'' and ``orange'' and to the classes ``Kleenex box'' and ``cereal box'' \aedit{, whereas the other dataset contains objects that have different shapes.}   

Furthermore, we test the robustness of RCFusion, trained on both the RGB and depth features, in real-world scenarios where the attacker can physically tamper with objects. To this end, we explore the effectiveness of adversarial patches computed to target the RGB part solely, and we compare its robustness with the one of ResNet-18. We aim to understand to which extent using a system that also leverages depth features may help with respect to employing a simpler ResNet-18.
To do so, we create adversarial patches against RCFusion and ResNet-18 and depict the collected results in Fig.~\ref{fig:adversarial_patch}, considering both the RGB-D Object and OCID datasets.

From the empirical results, we highlight that the influence of the adversarial patches on the robustness of both RCFusion and ResNet-18 are very similar, with RCFusion being slightly more accurate.
Such minimal discrepancy might be caused by (i) the additional complexity that is included inside the architecture of RCFusion and (ii) the additional information provided by the depth information.
However, this advantage still decreases with the increase in the size of the adversarial patch.
Therefore, it is easy to conclude that the robustness of RCFusion is almost as vulnerable to adversarial patches as ResNet-18. \textit{This means that using RCFusion, which is a more complex system than ResNet-18, which leverage also the depth, does not provide relevant advantages in terms of security.}

\begin{figure}[t]
    \centering
    \subfloat[]
    {\includegraphics[trim=0 1.2cm 4cm 1.5cm, clip, width=.232\textwidth]{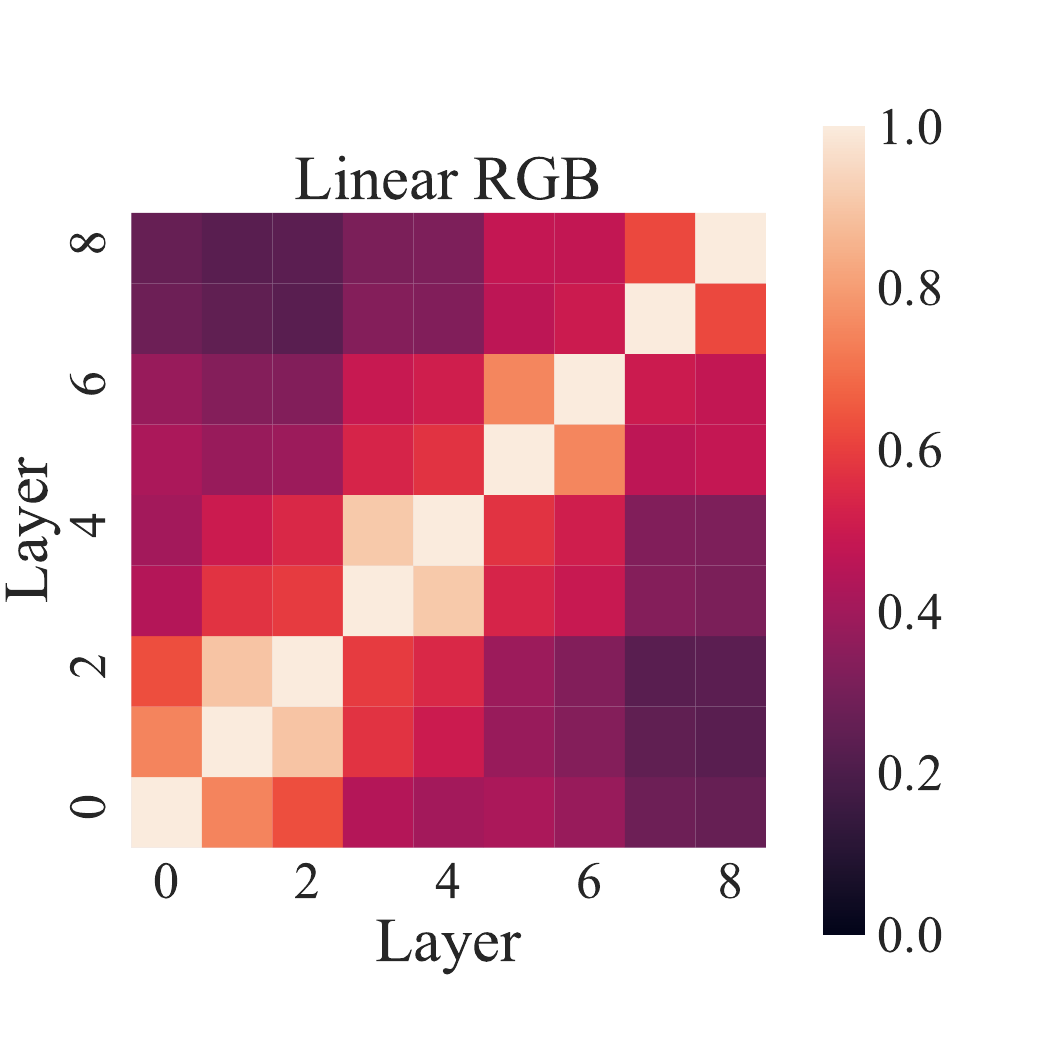}
	\includegraphics[trim=0 1.2cm 4cm 1.5cm, clip, width=.232\textwidth]{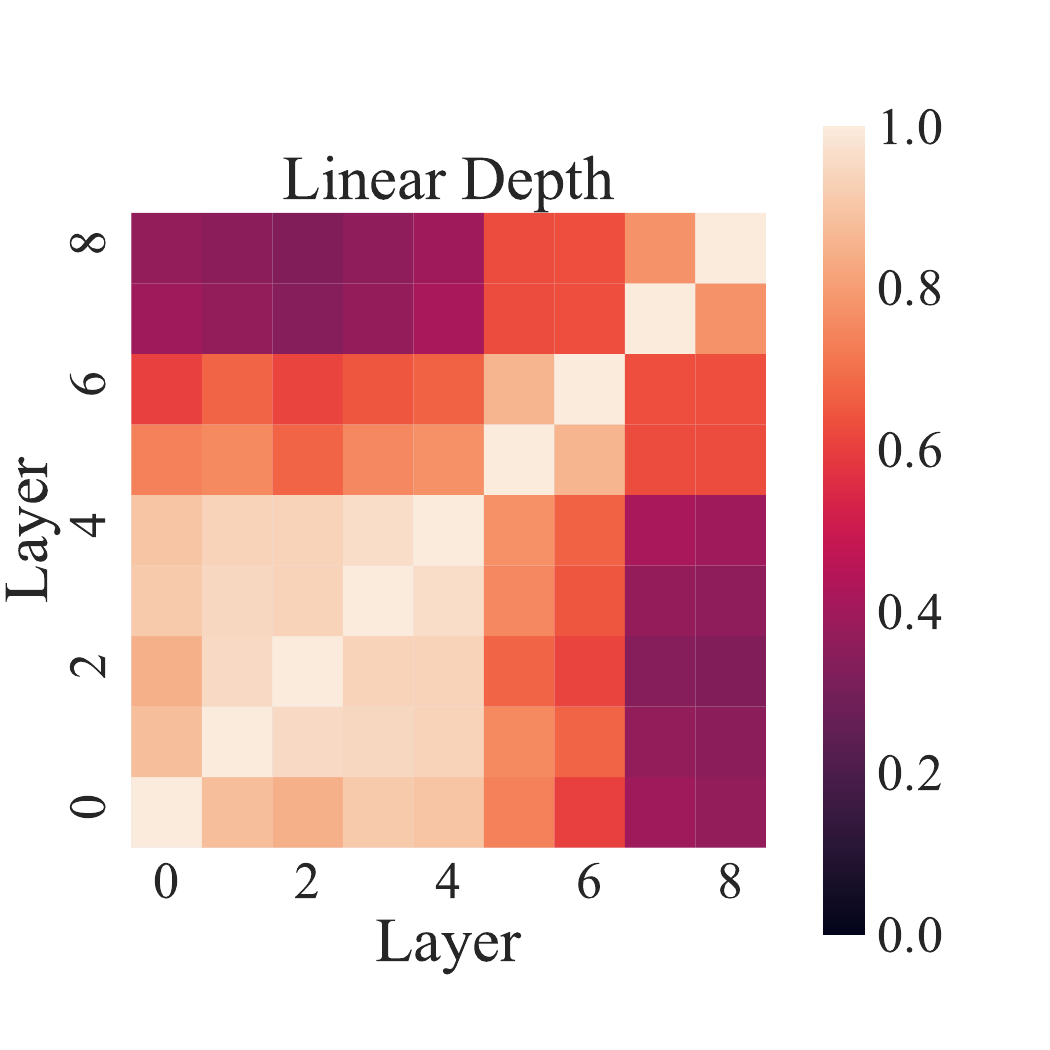}
    \includegraphics[trim=0 1.2cm 4cm 1.5cm, clip, width=.232\textwidth]{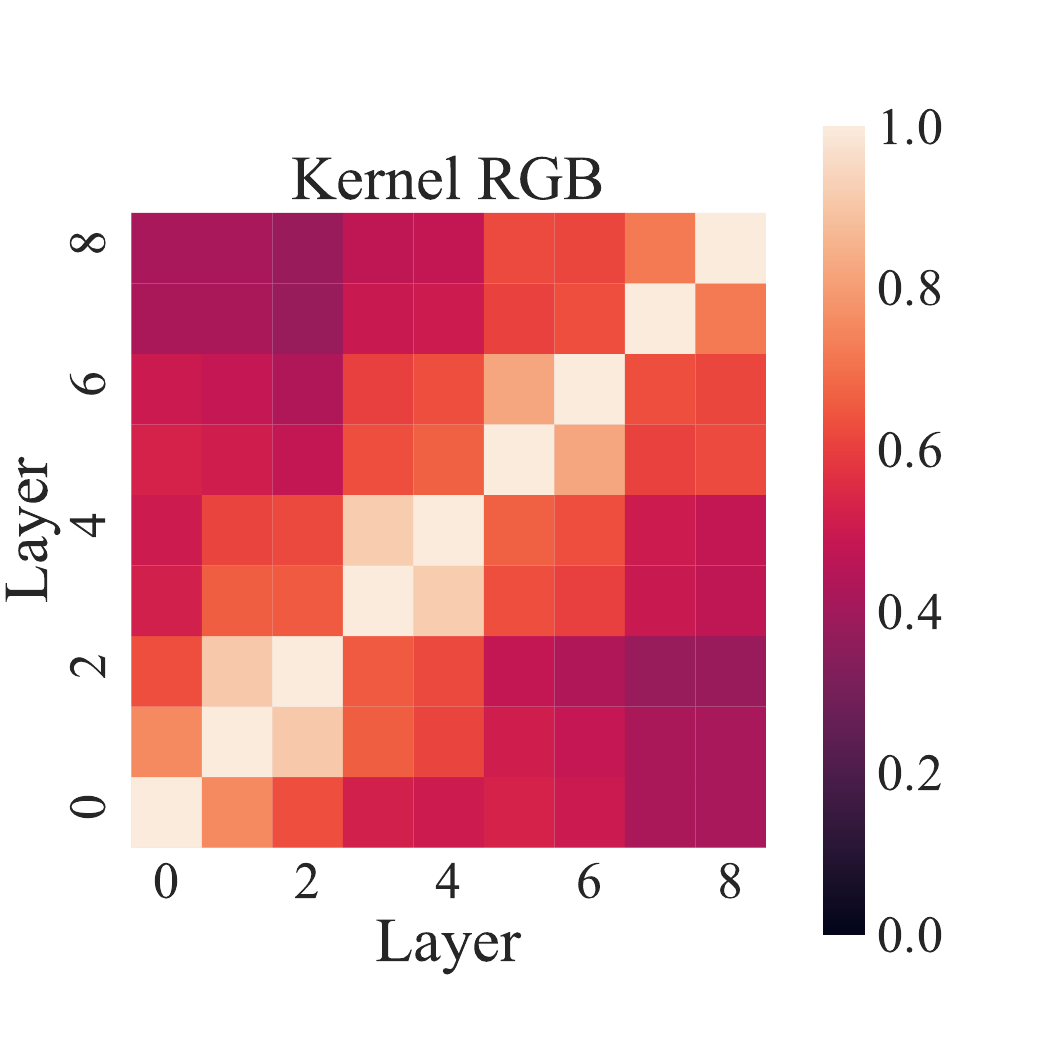}
	\includegraphics[trim=0 1.2cm 0 1.5cm, clip, width=.30\textwidth]{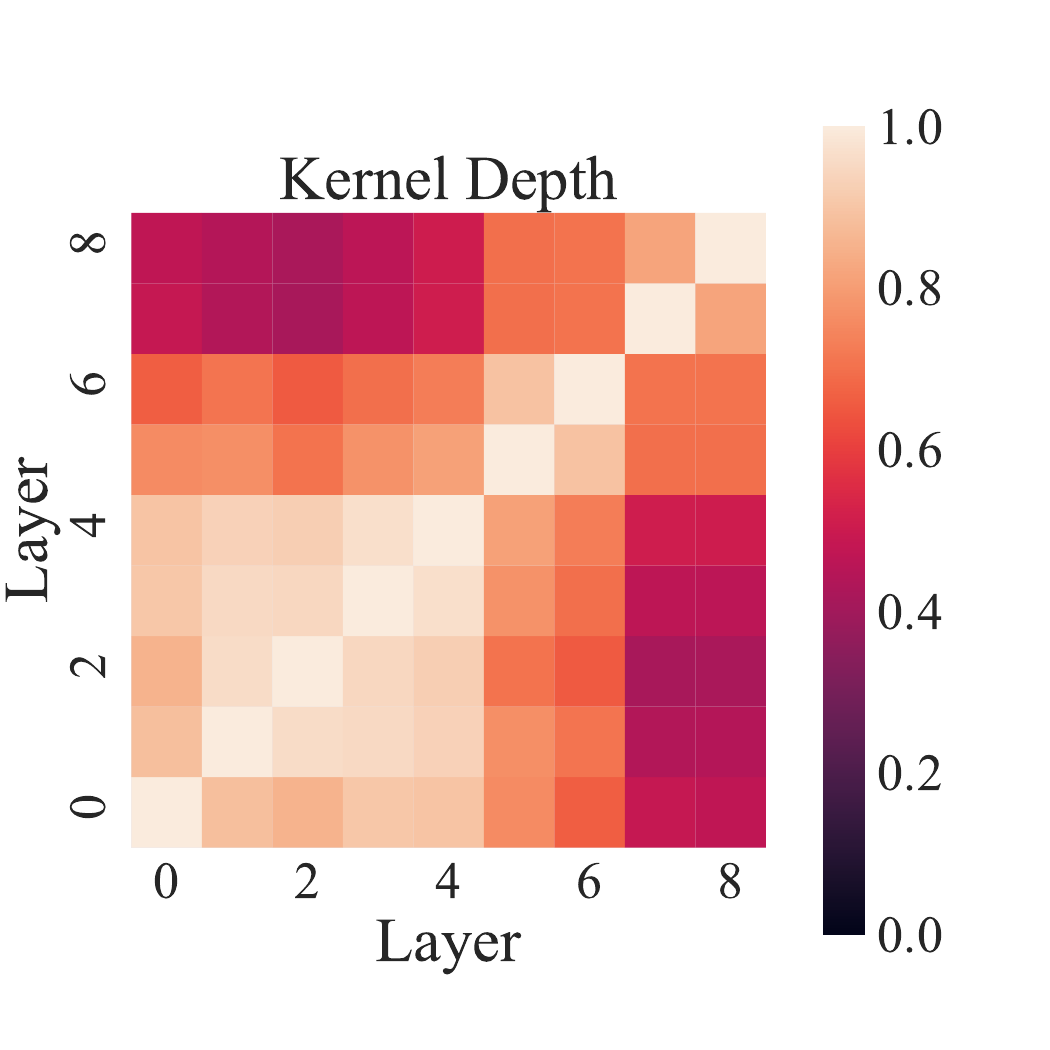}
    \label{fig:heatmap:a}}\hfill
    \subfloat[]
    {\includegraphics[trim=0 1.2cm 4cm 1.5cm, clip, width=.232\textwidth]{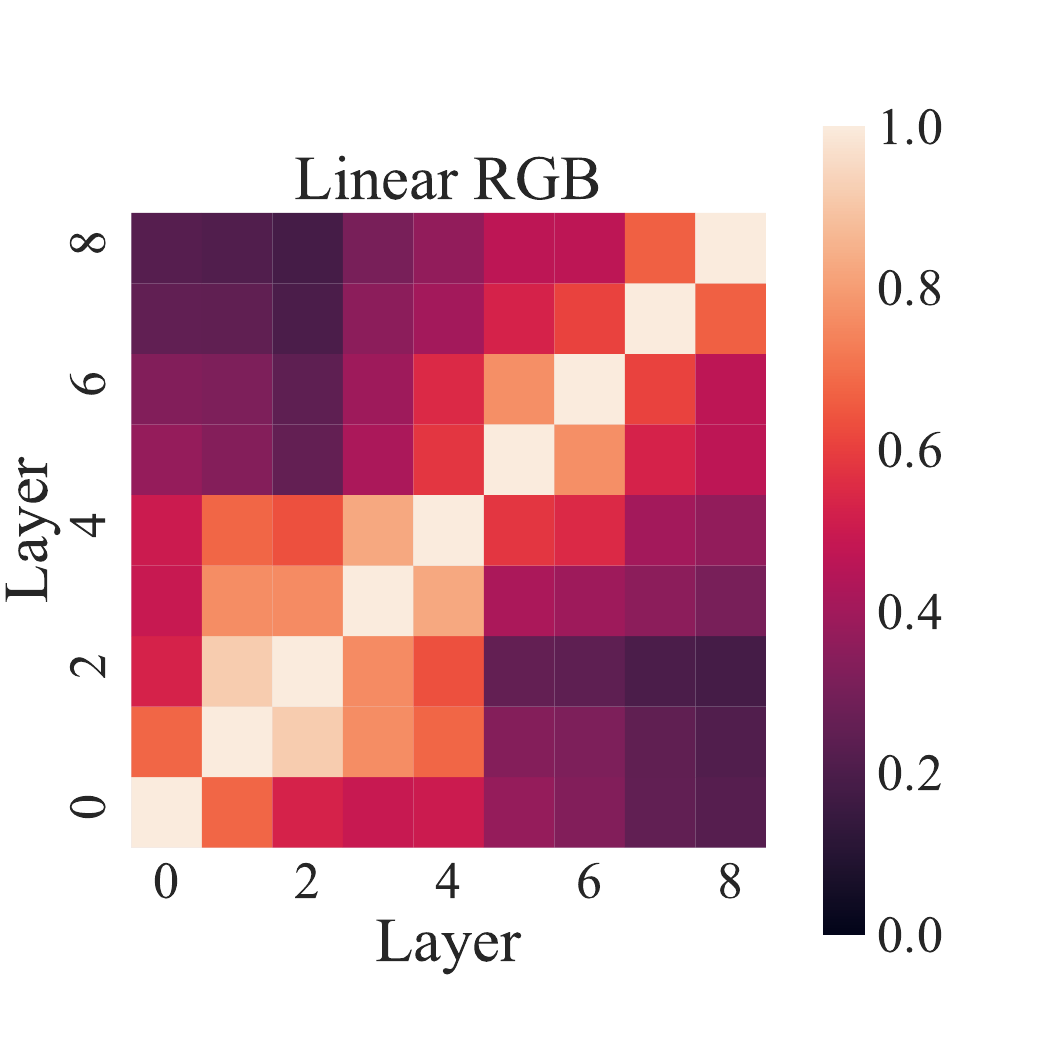}
	\includegraphics[trim=0 1.2cm 4cm 1.5cm, clip, width=.232\textwidth]{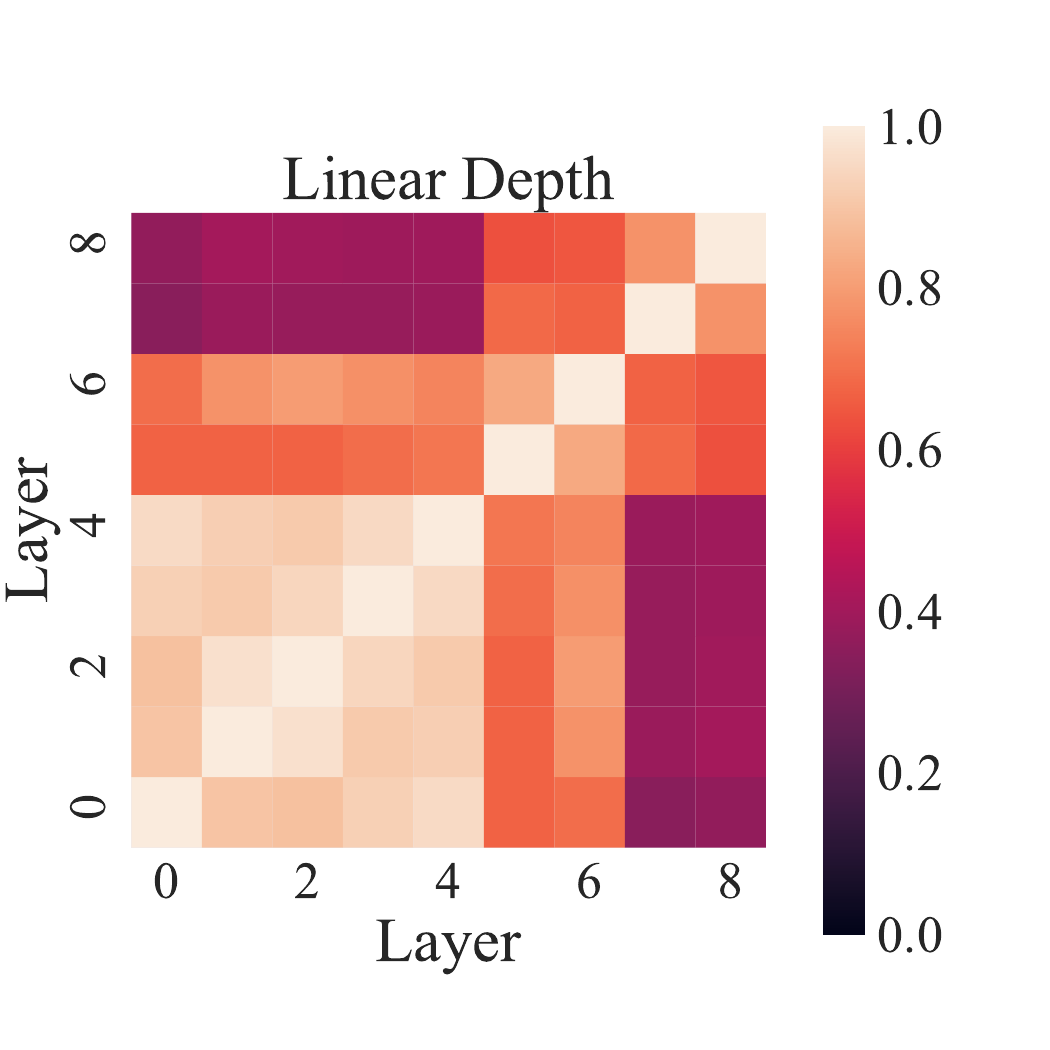}
    \includegraphics[trim=0 1.2cm 4cm 1.5cm, clip, width=.232\textwidth]{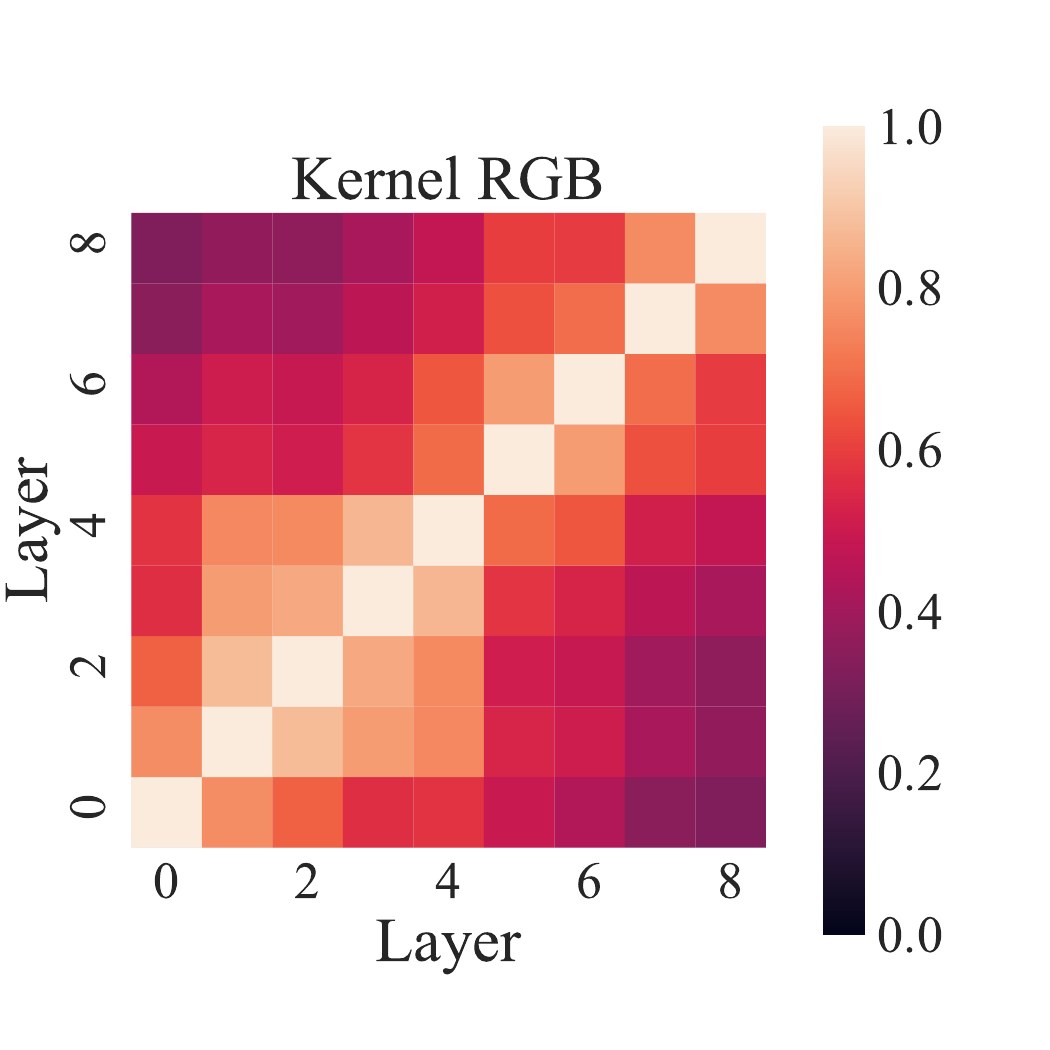}
	\includegraphics[trim=0 1.2cm 0 1.5cm, clip, width=.30\textwidth]{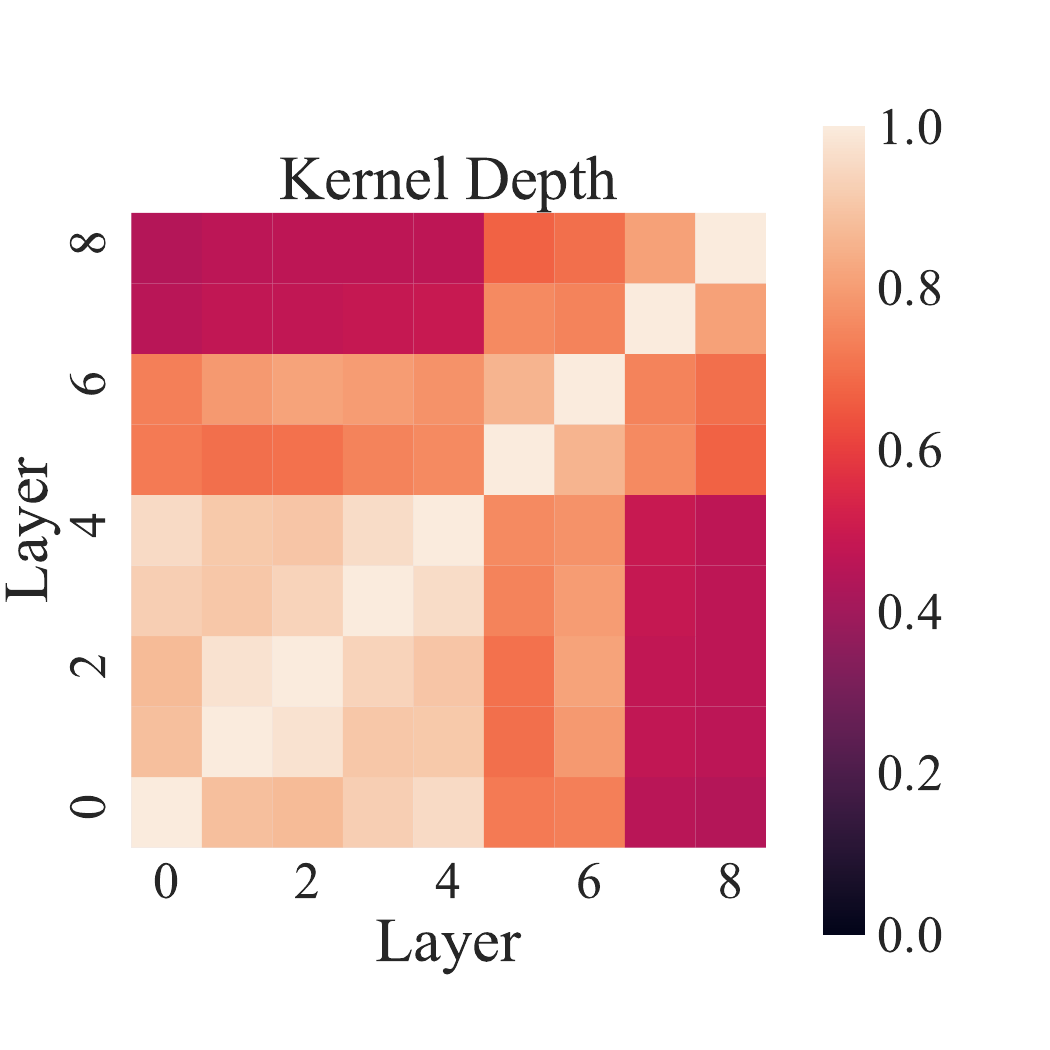}
    \label{fig:heatmap:b}}
    \caption{The CKA similarity heatmaps generated based on the RGB and depth information by applying the linear (first two) and RBF (last two) kernels, where (a) is the results of RGB-D Object Dataset, (b) is the results of OCID. }
    \label{fig:heatmap}
\end{figure}

\myparagraph{Interpretation of RCFusion Vulnerability.}
In the following, we analyze the internal representations in RCFusion to explain why it is vulnerable to adversarial attacks. To this end, we use the CKA measure, seen in Section~\ref{sec:robustness}, to evaluate the similarity of features abstracted by RCFusion from the RGB and depth part separately and present the CKA similarity heatmaps in Fig.~\ref{fig:heatmap}. We can see: (i) the linear and RBF kernels give similar results on the RGB-D Object Dataset and OCID. This conclusion is consistent with that presented in \citep{kornblith2019similarity}; (ii) the heatmaps generated on the depth information tend to show a distinctive block structure\citep{nguyen2020wide} (seen as a yellow square on the heatmap) than the RGB information. We conjecture this is because \textit{the information learned by the network trained on depth is more redundant} (there is less information to learn). Therefore, the trained classifier turns out to be smoother and, thus, more robust to input perturbations. 
\smallskip 

\begin{figure}[t]
\centering
    \includegraphics[width=0.45\textwidth]{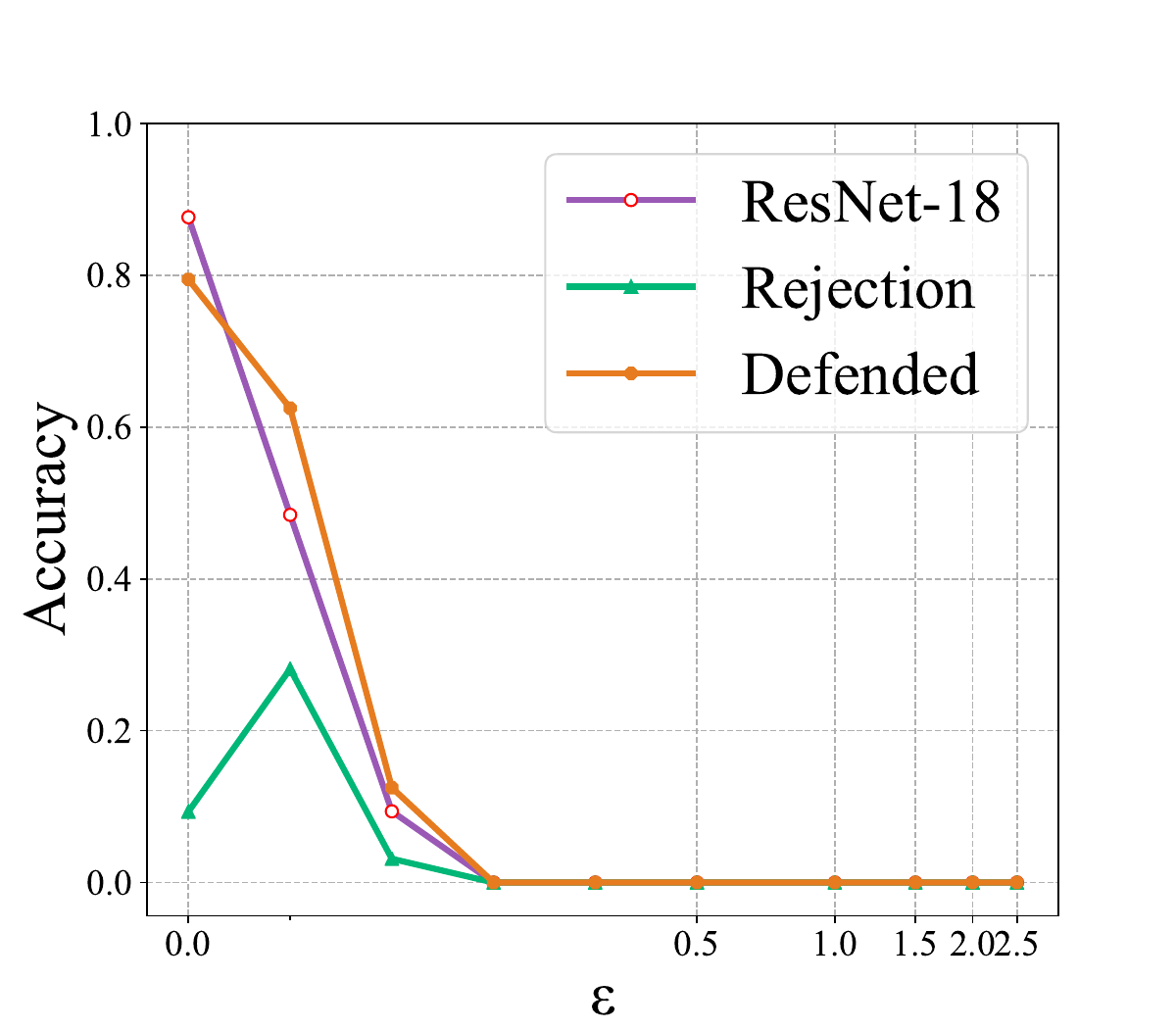}
    \includegraphics[width=0.45\textwidth]{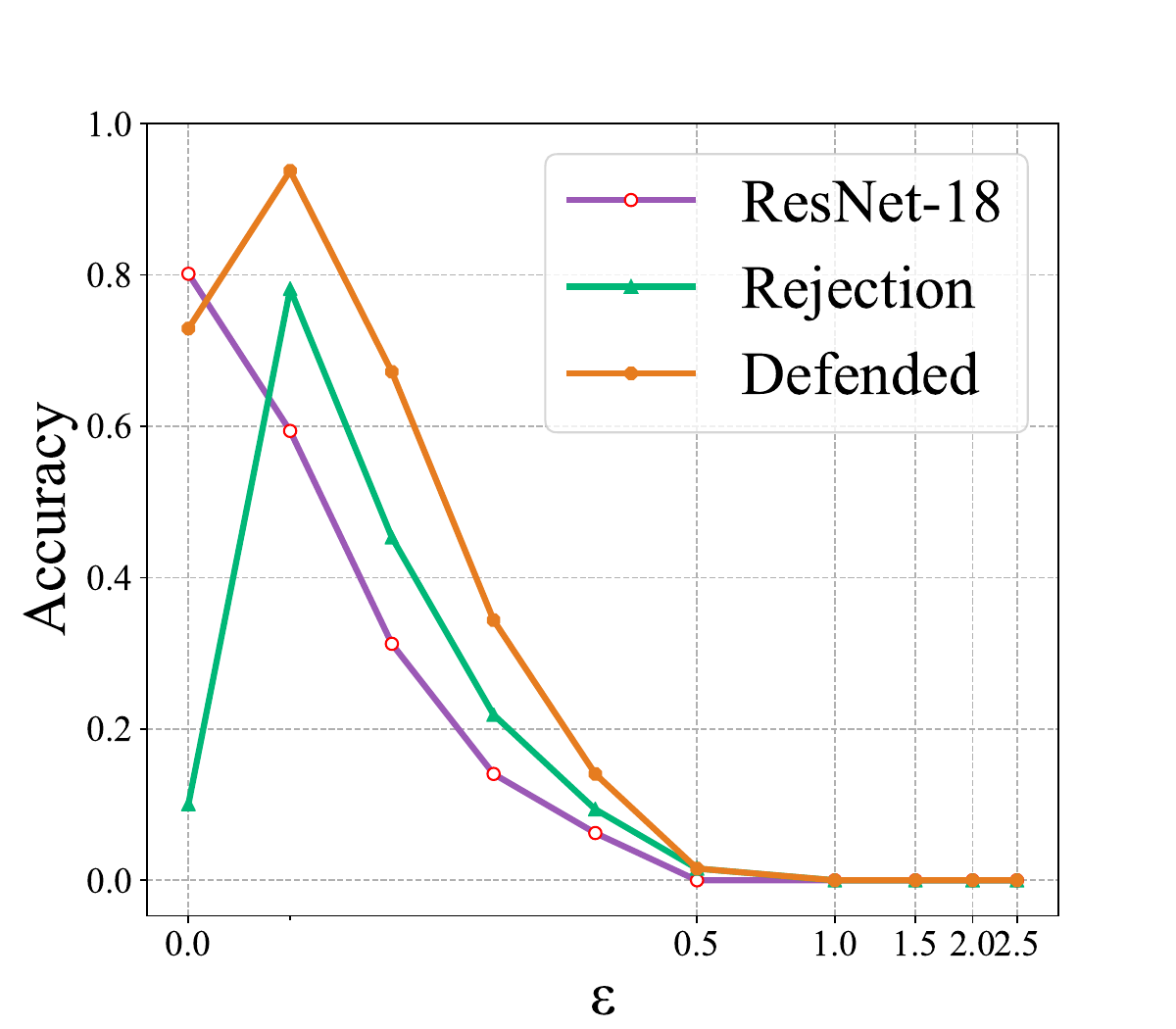}
    \hfill
    \includegraphics[width=0.45\textwidth]{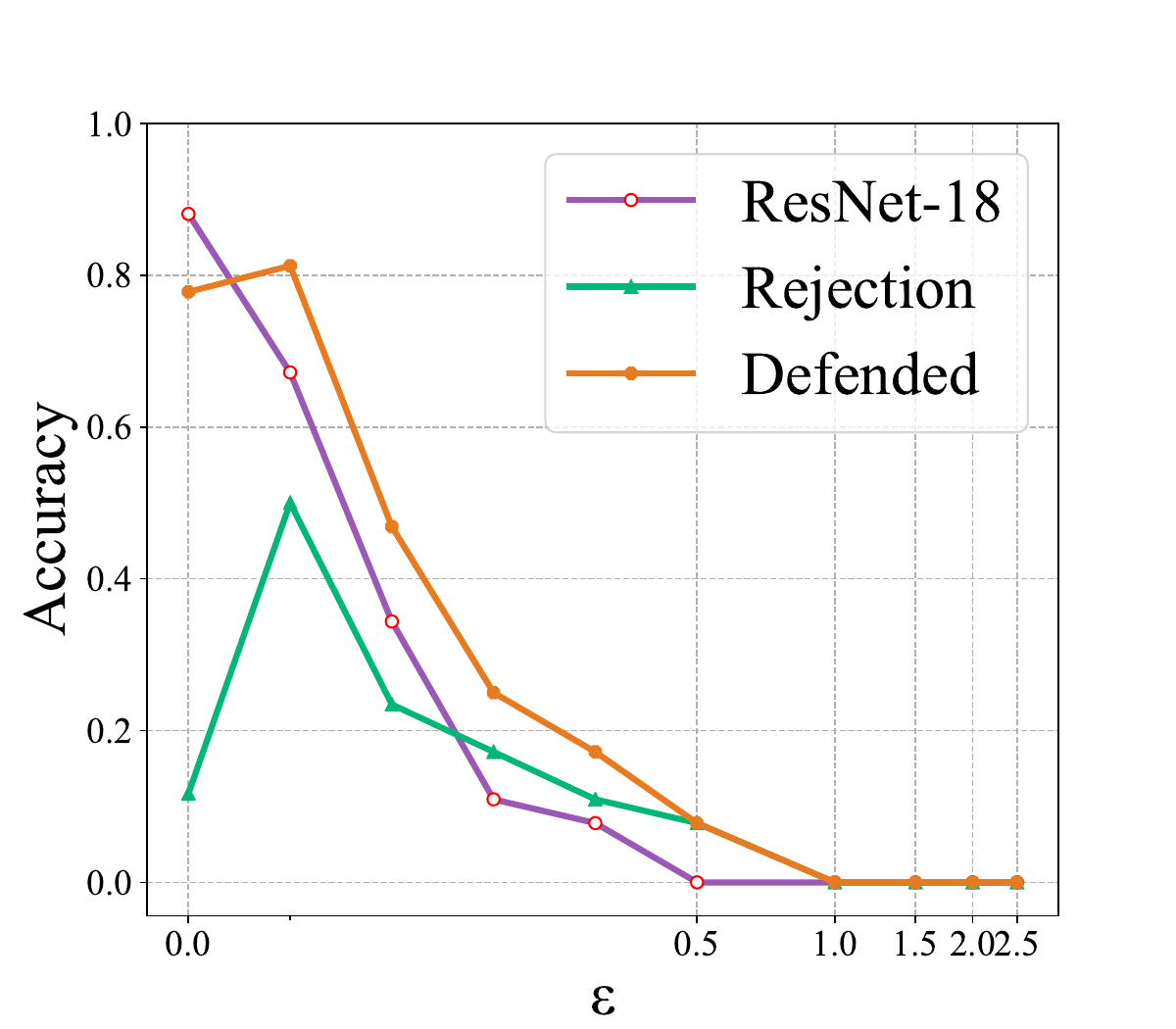}
    \includegraphics[width=0.45\textwidth]{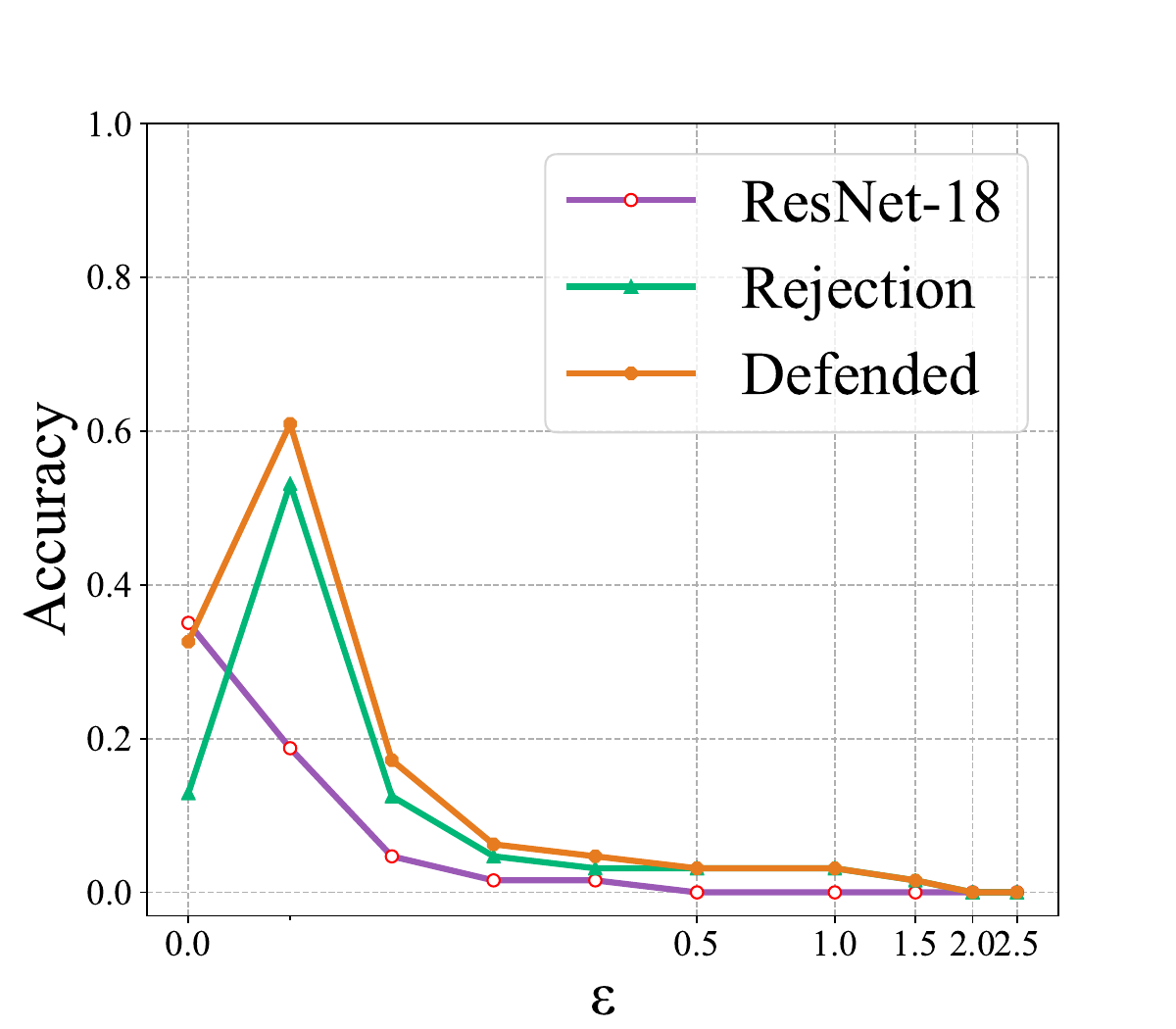}
    \caption{The robustness evaluation curve computed by modifying the full image of the input of ResNet-18 trained on the RGB (left) and depth (right) separately for the RGB-D Object Dataset (top-row) and OCID Dataset (bottom-row).}    
\label{fig:rob_resnet18}
\end{figure}
\begin{figure}[t]
\centering
    {\includegraphics[width=0.45\textwidth]{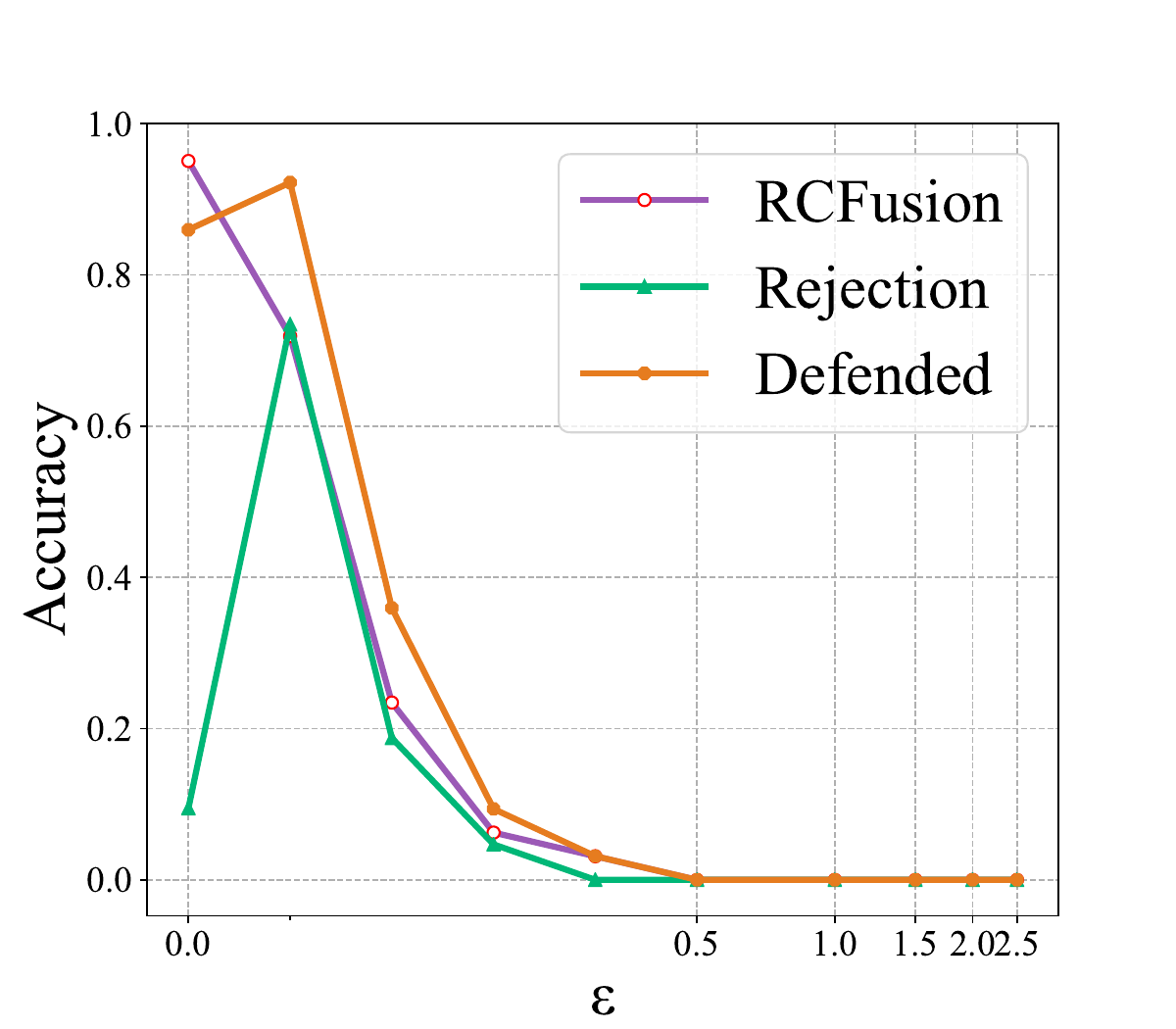}
        \includegraphics[width=0.45\textwidth]{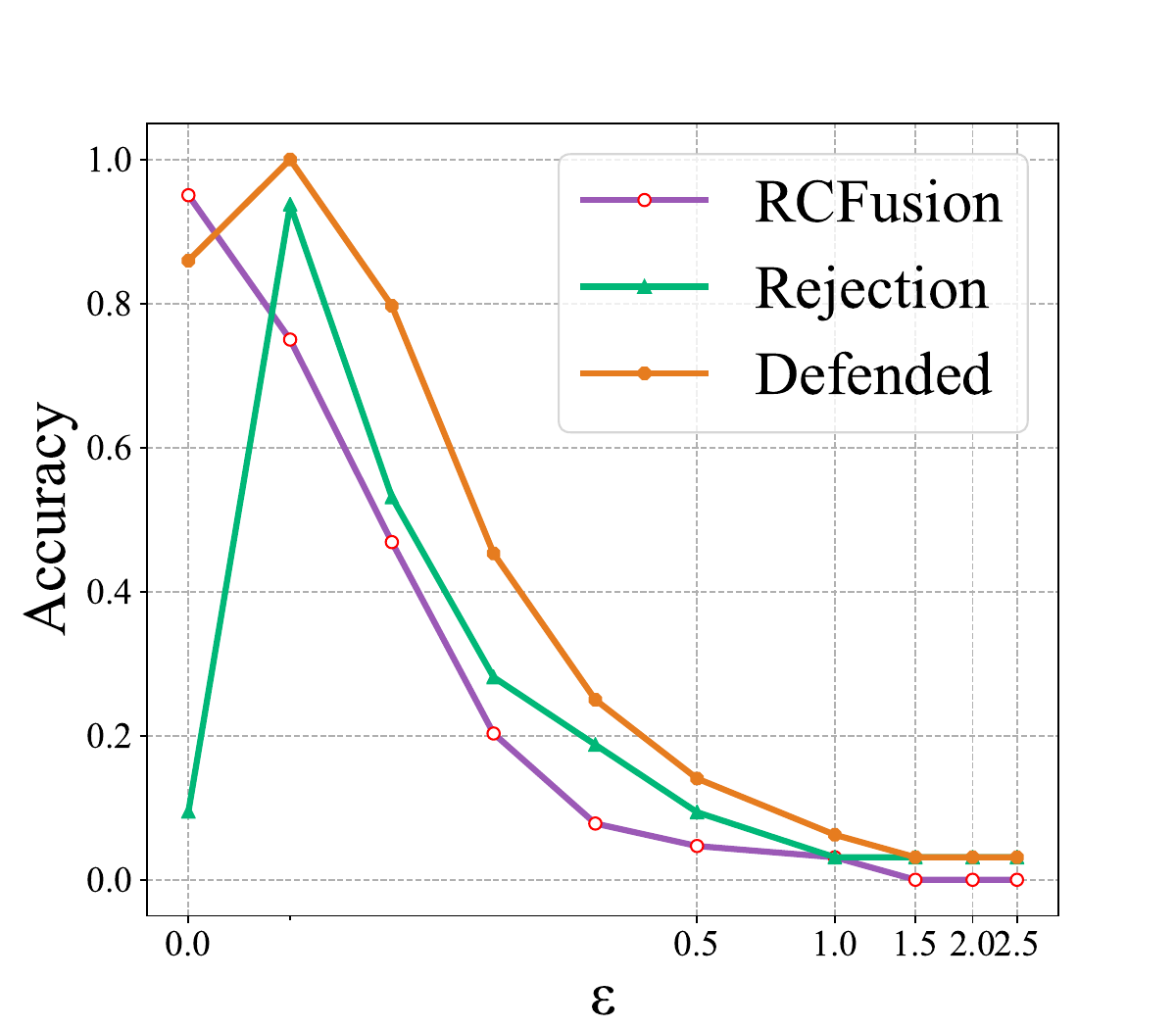}}
    
    {\includegraphics[width=0.45\textwidth]{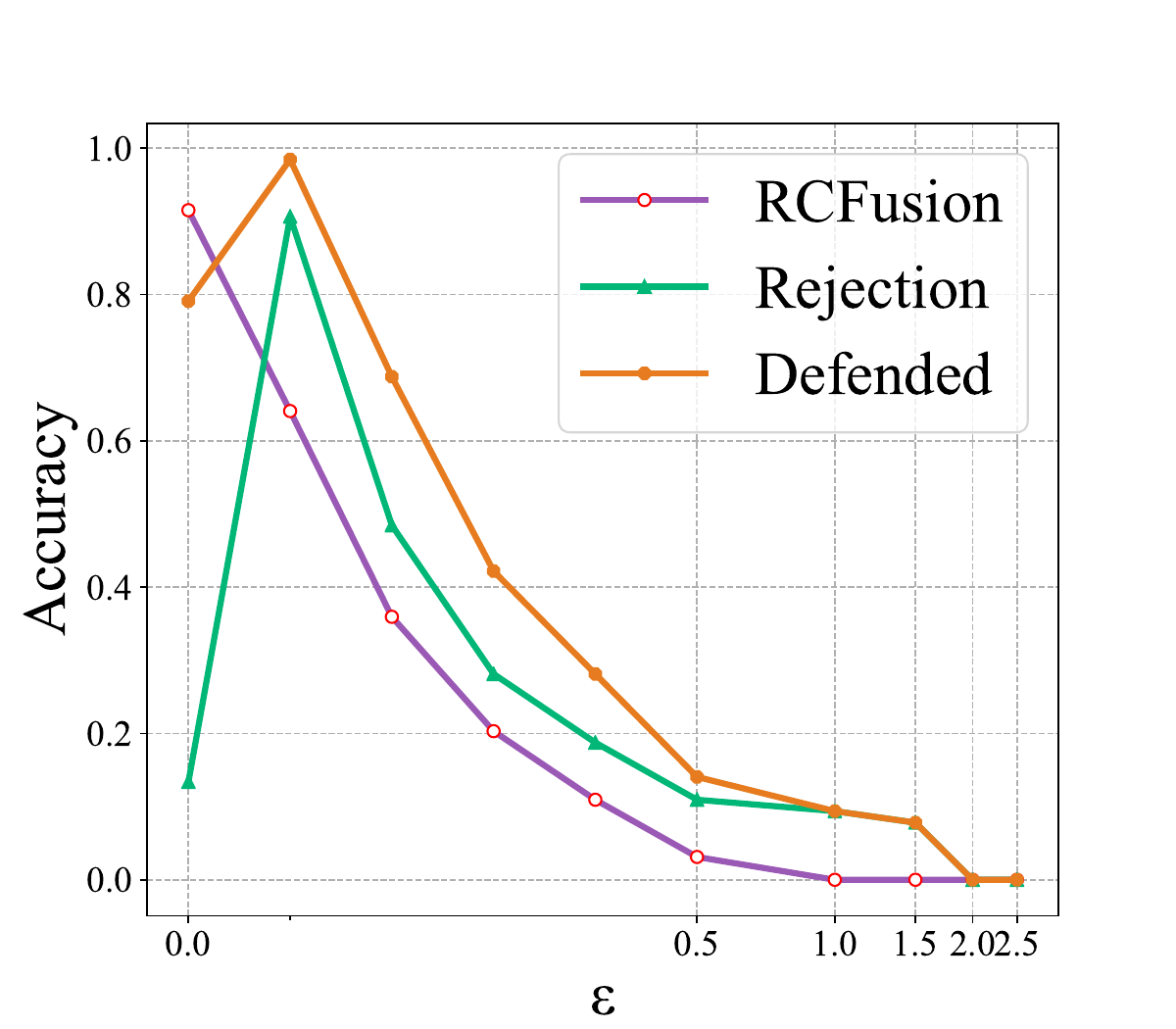}
    \includegraphics[width=0.45\textwidth]{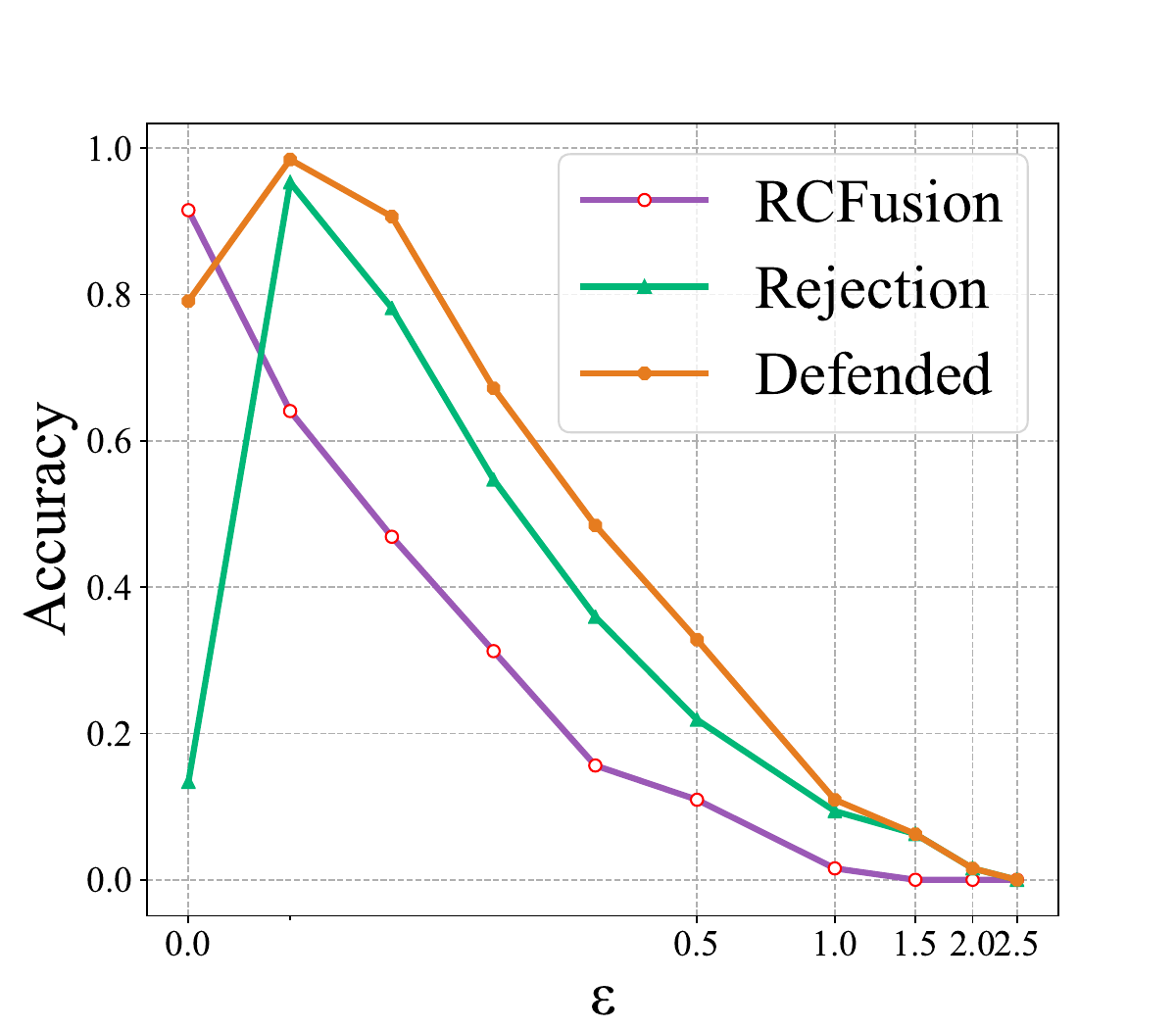}}
    \caption{The robustness evaluation curve is computed by modifying the full image of the RGB-D (left) and RGB (right) channel of RCFusion constructed by ResNet-18 on the RGB-D Object Dataset (top-row) and OCID (bottom-row), when the step size is 0.05.}
\label{fig:rob_rcfusion_all}
\end{figure}
\begin{figure}[t]
\centering
    {\includegraphics[width=0.45\textwidth]{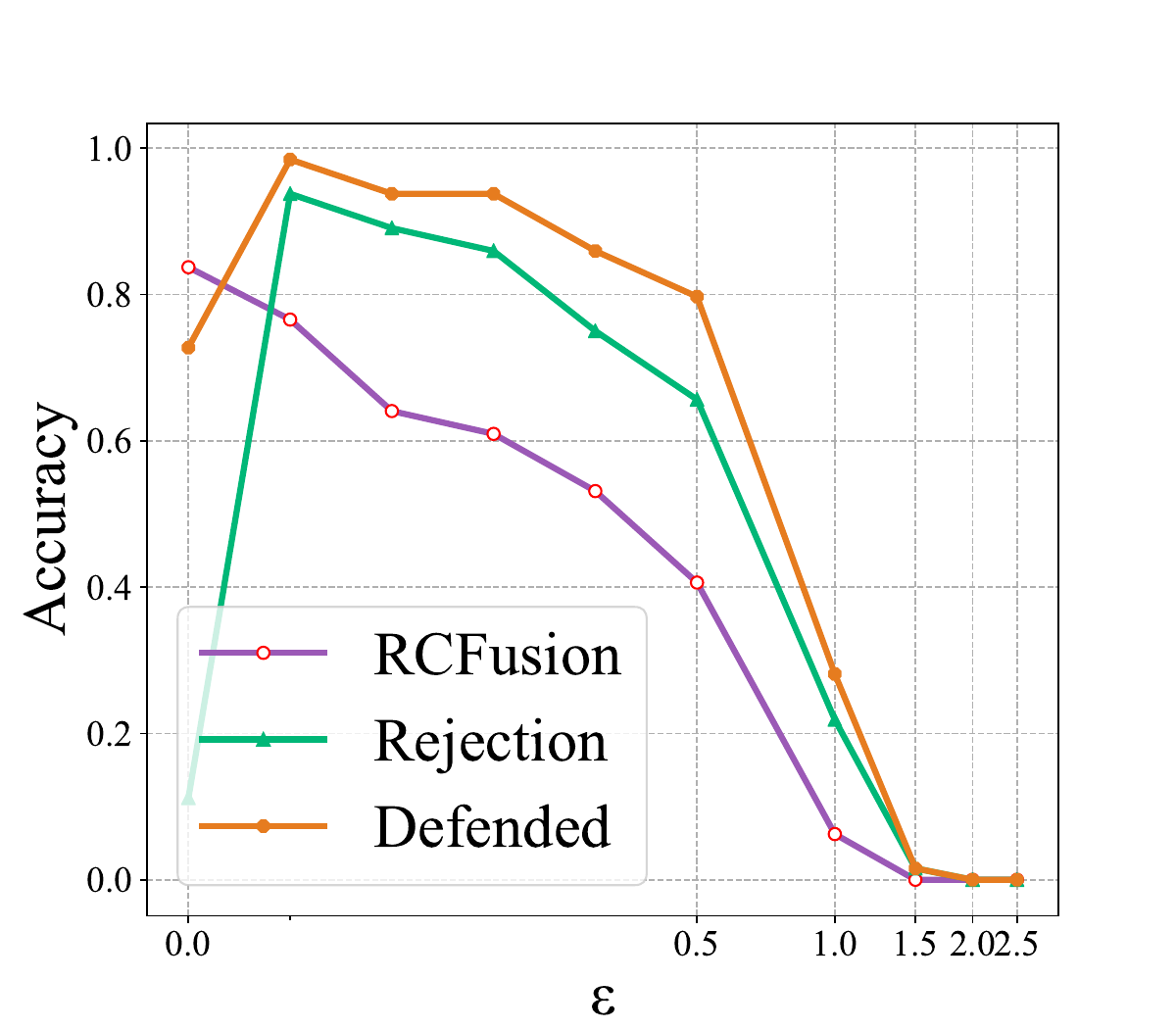}
        \includegraphics[width=0.45\textwidth]{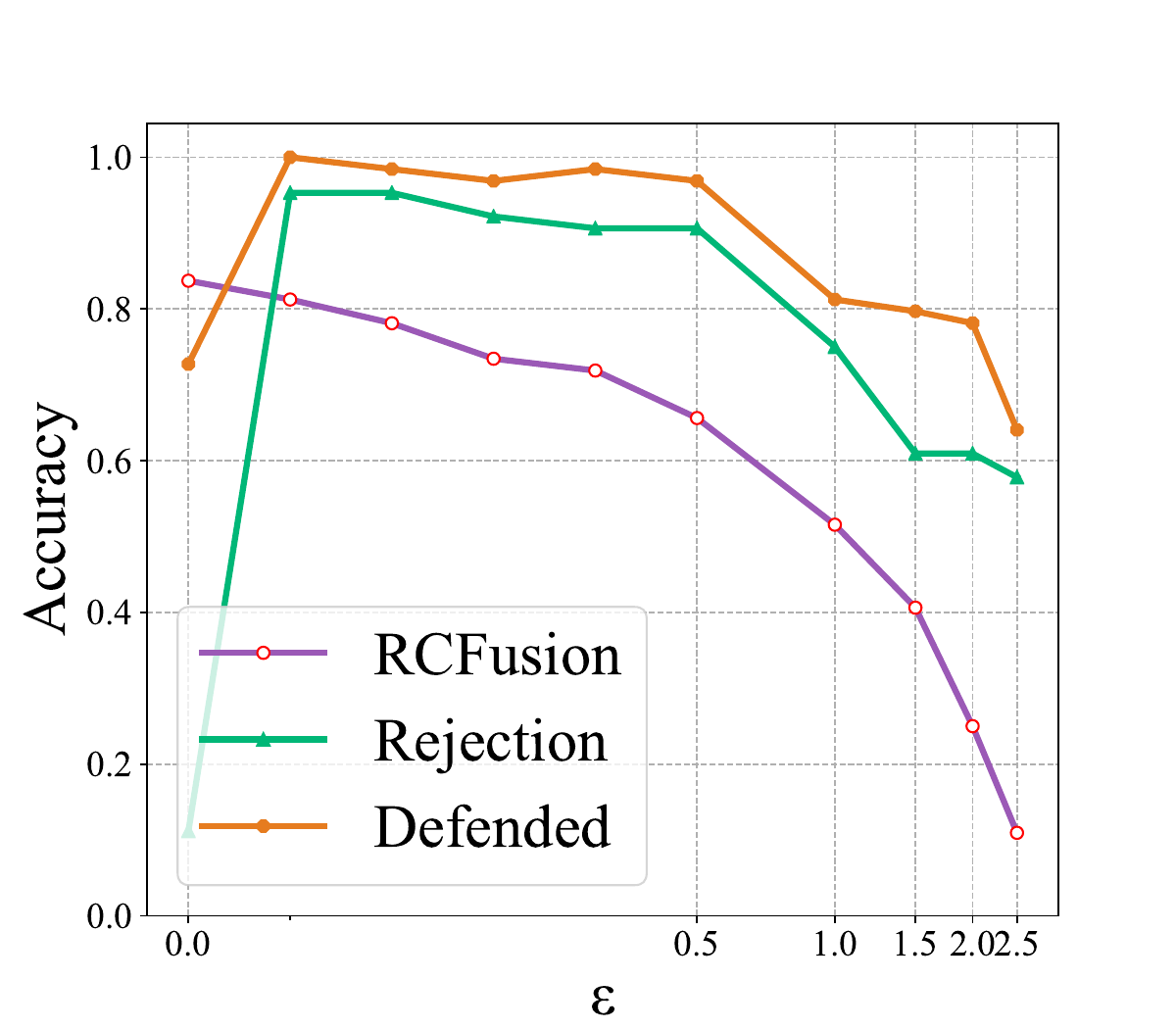}}
    {\includegraphics[width=0.45\textwidth]{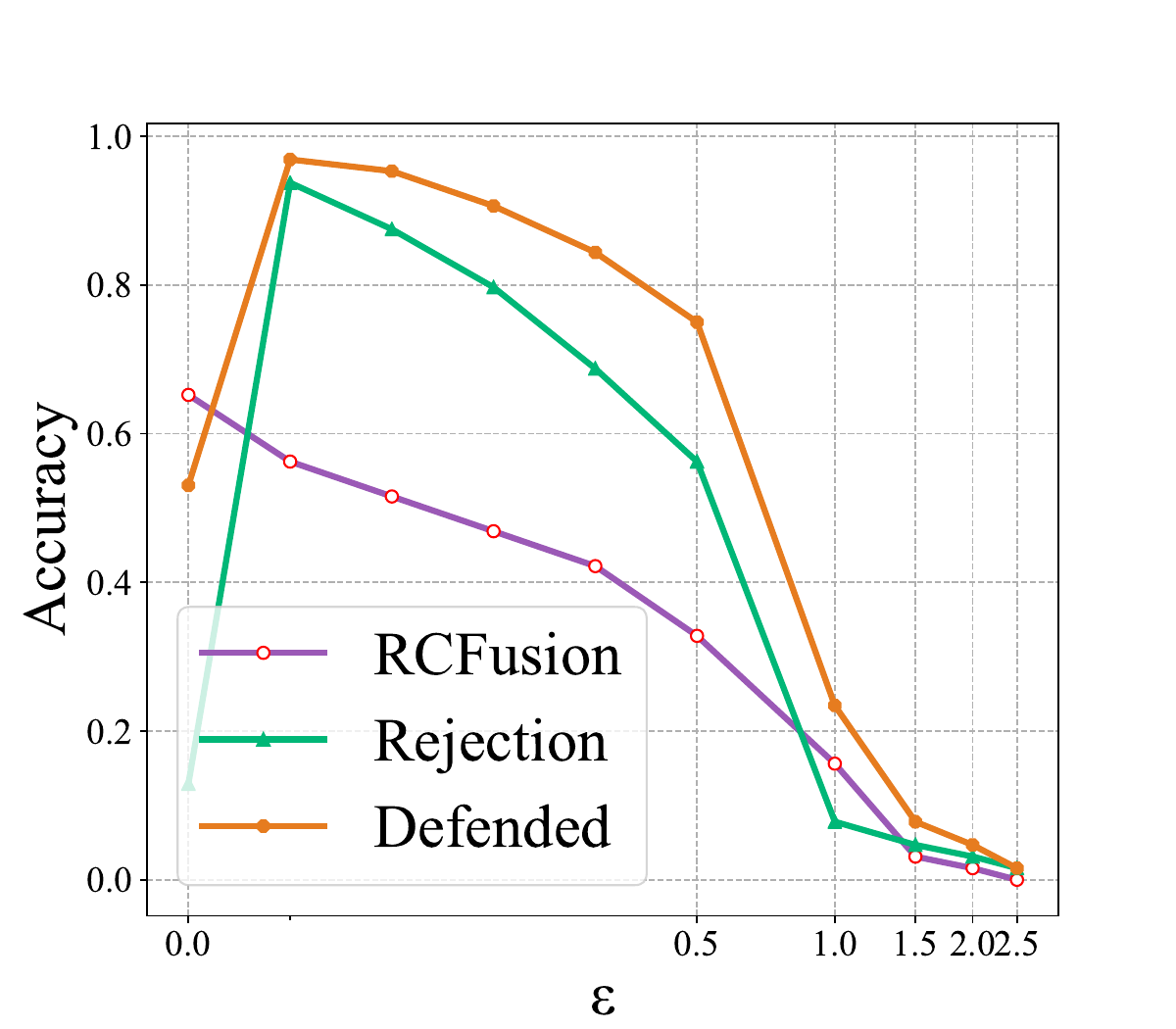}
    \includegraphics[width=0.45\textwidth]{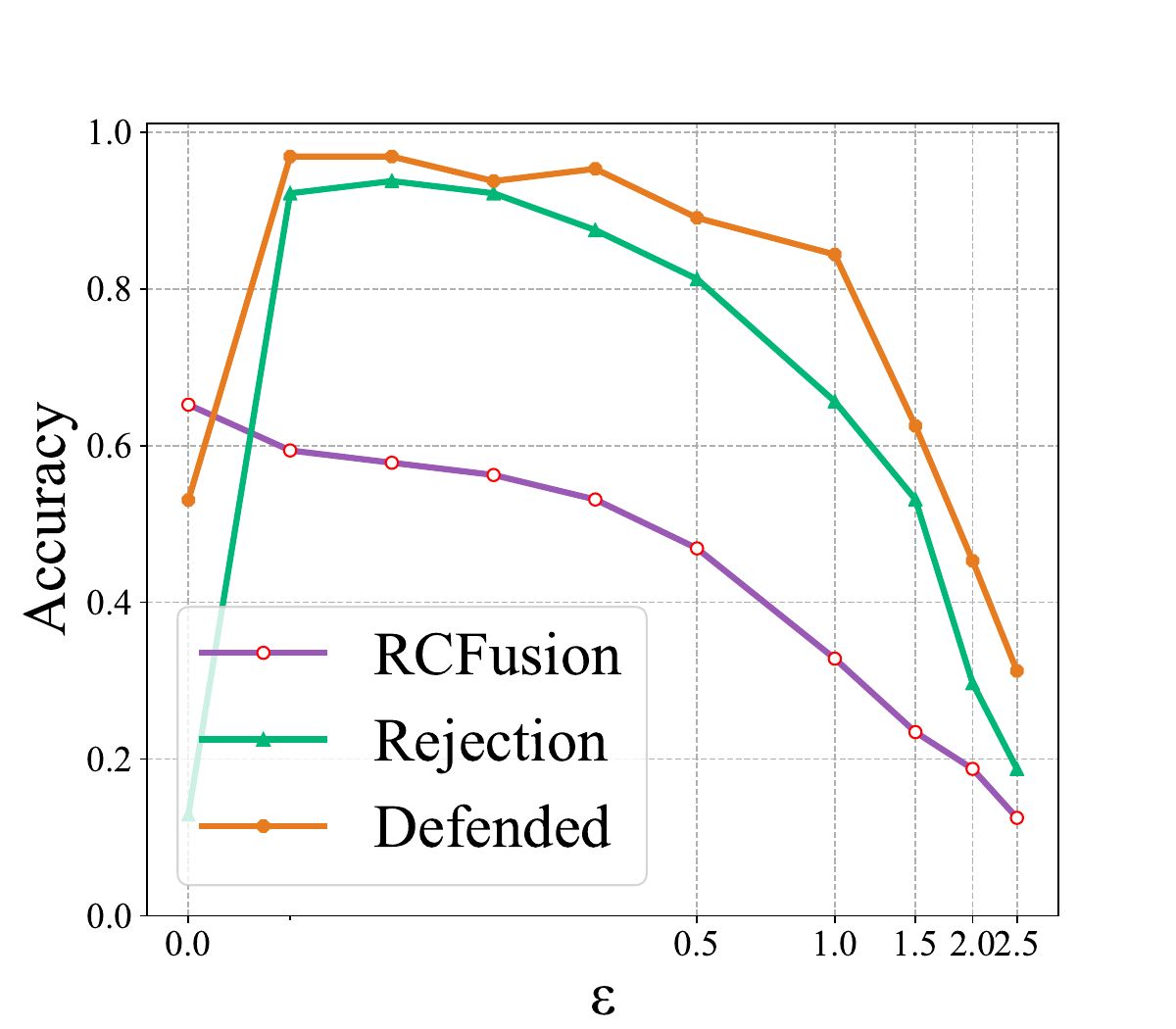}}
    \caption{\yzzedit{The robustness evaluation curve is computed by modifying the full image of the RGB-D (left) and RGB (right) channel of RCFusion constructed by AlexNet on the RGB-D Object Dataset (top-row) and OCID (bottom-row), when the step size is 0.05.}}
\label{fig:rob_rcfusion_alexnet_all}
\end{figure}
\begin{figure}[h!]
\centering
    {\includegraphics[width=0.48\textwidth]{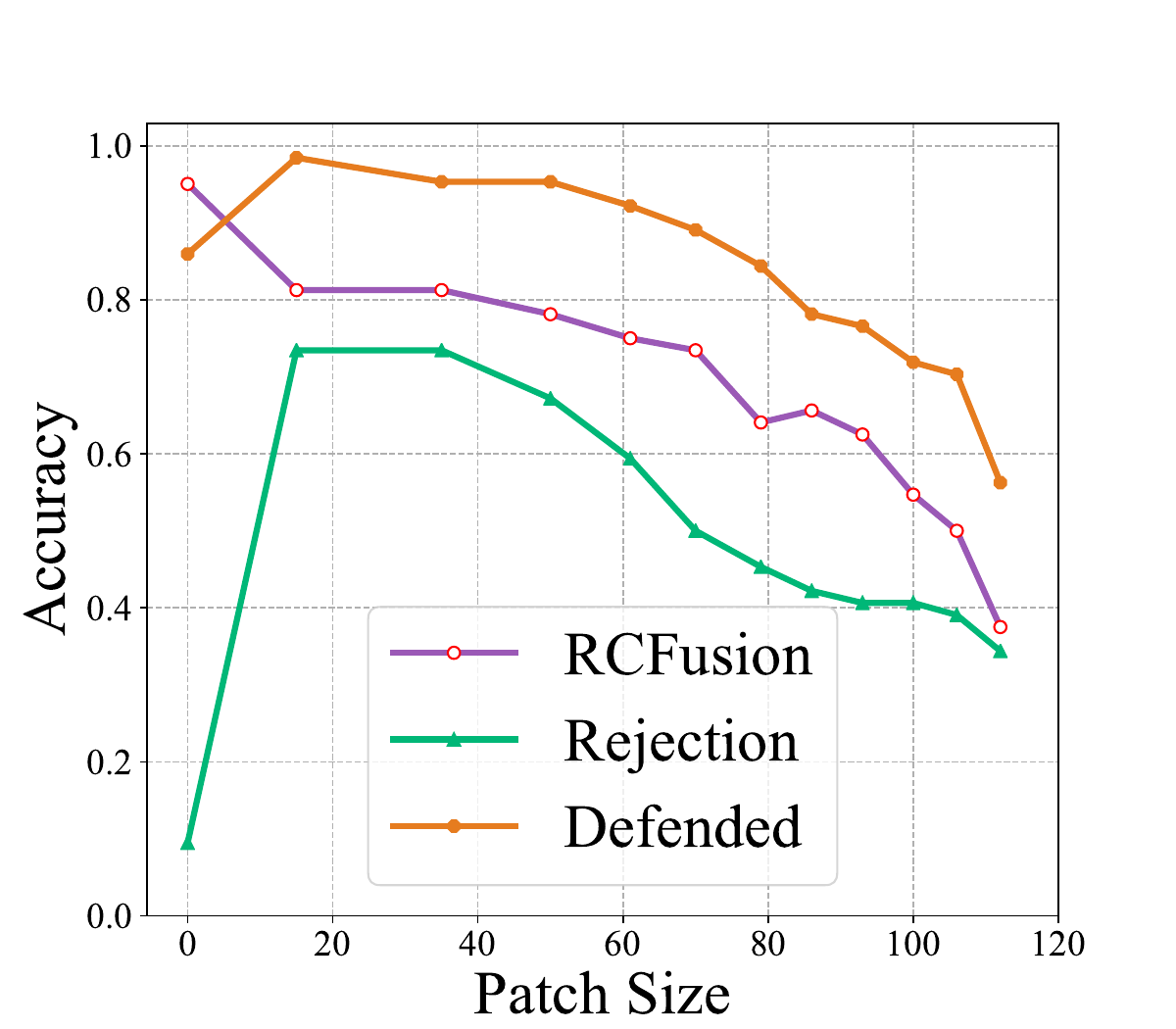}
    \includegraphics[width=0.48\textwidth]{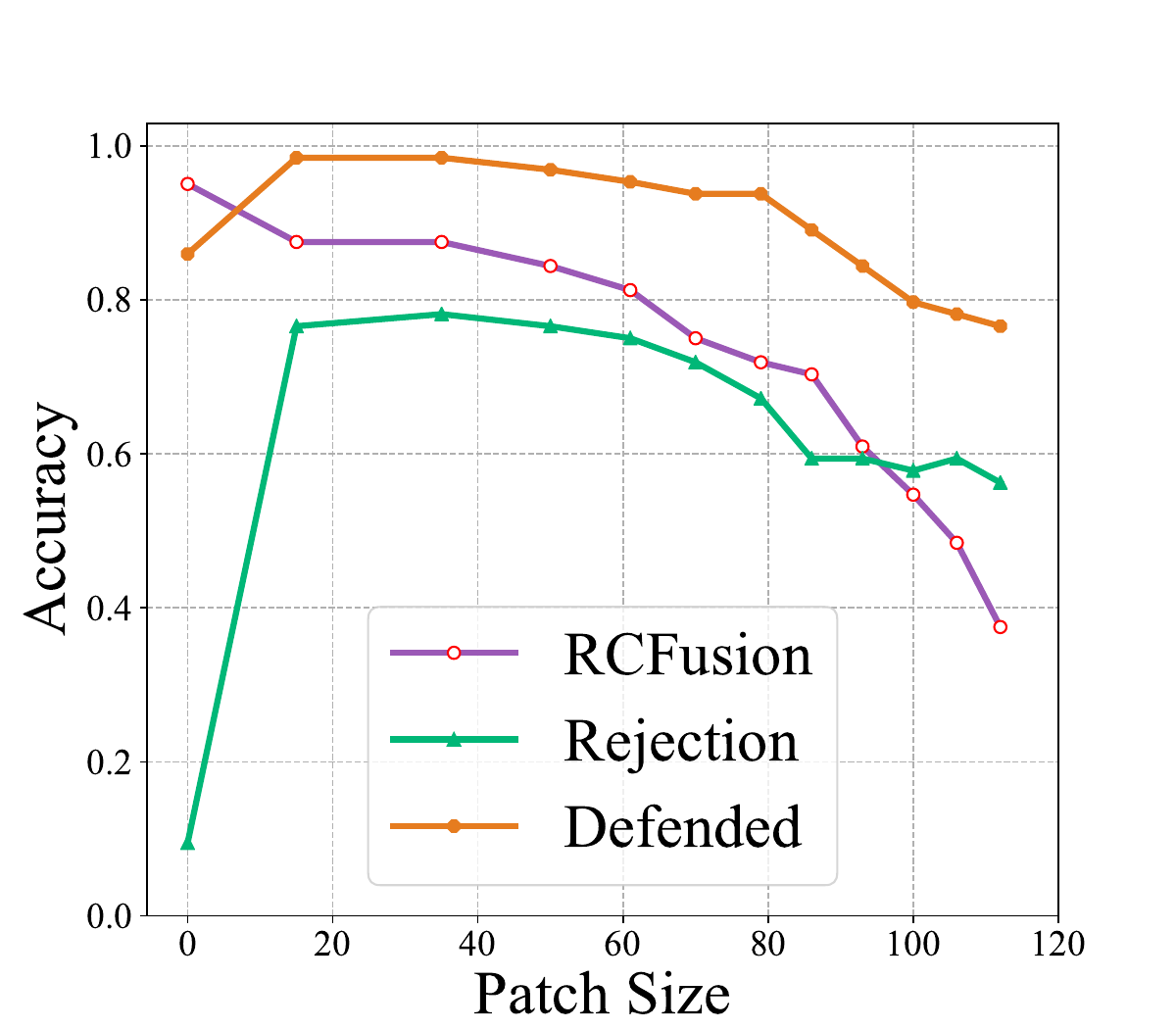}}
    {\includegraphics[width=0.48\textwidth]{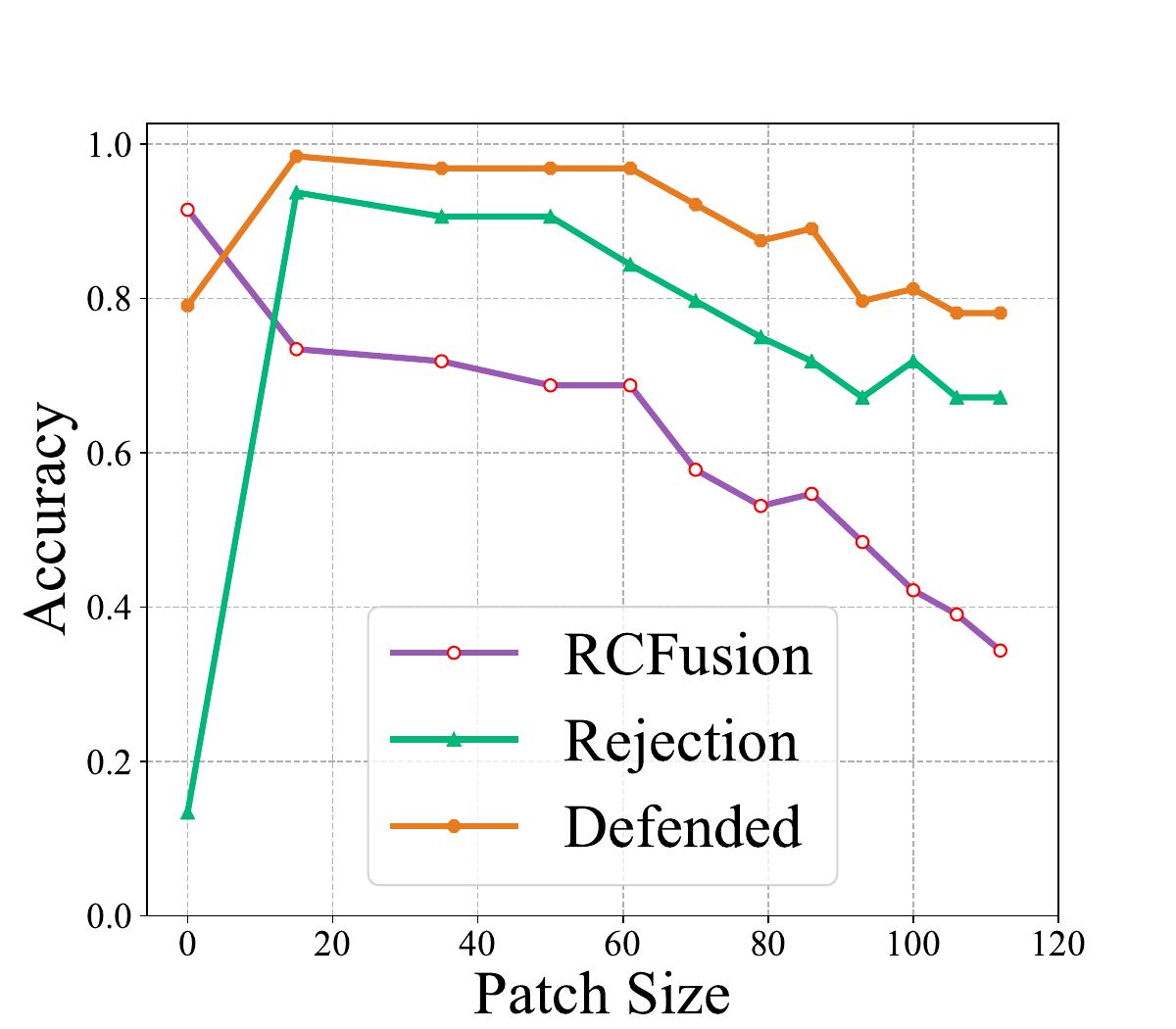}
    \includegraphics[width=0.48\textwidth]{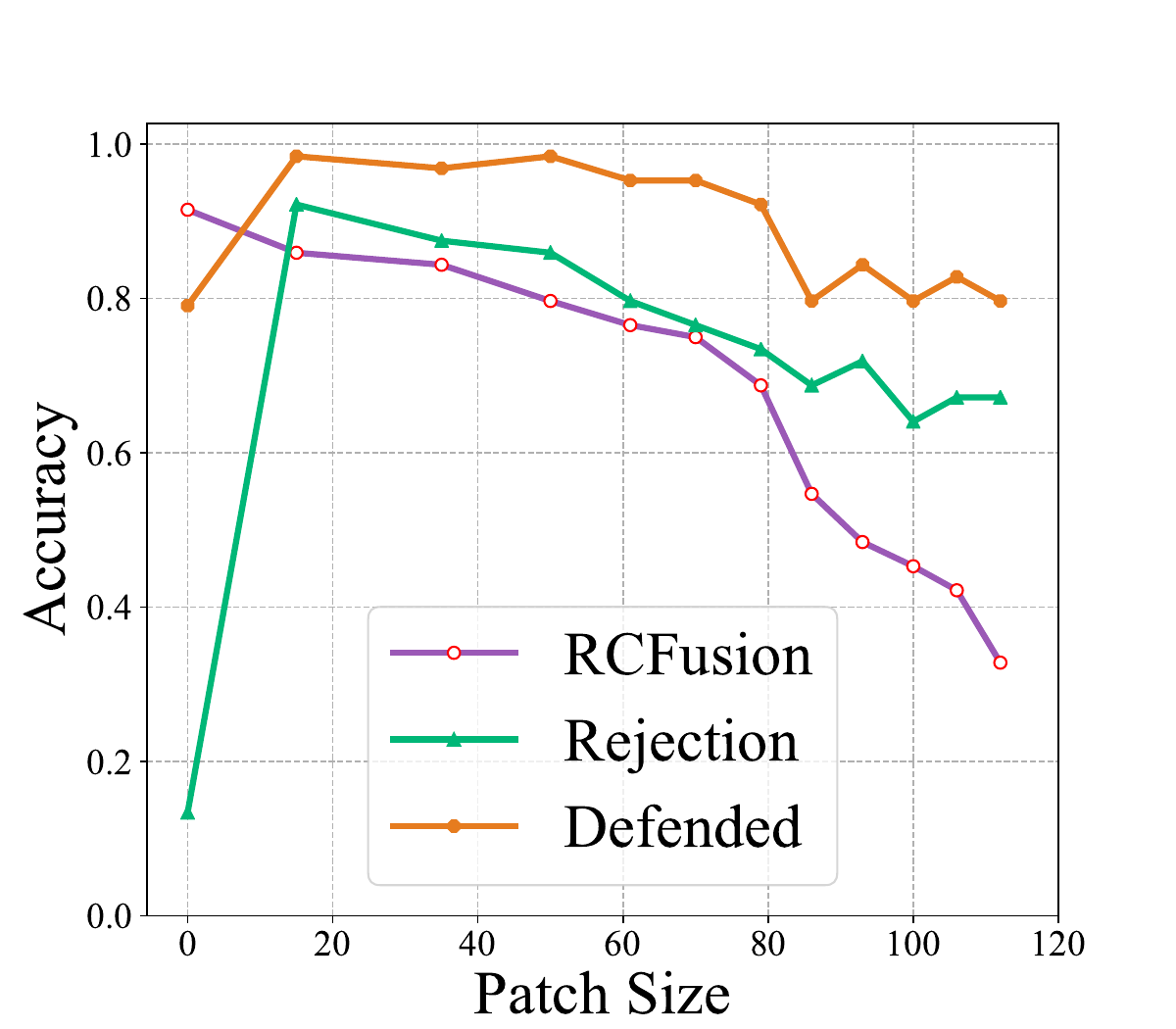}}
    \caption{The robustness evaluation curve is computed by modifying only a portion image of the RGB-D (left) and RGB (right) channel of RCFusion constructed by ResNet-18 on the RGB-D Object Dataset (top-row) and the OCID Dataset (bottom-row).}
\label{fig:rob_rcfusion_patch}
\end{figure}
\begin{figure}[h!]
\centering
    {\includegraphics[width=0.48\textwidth]{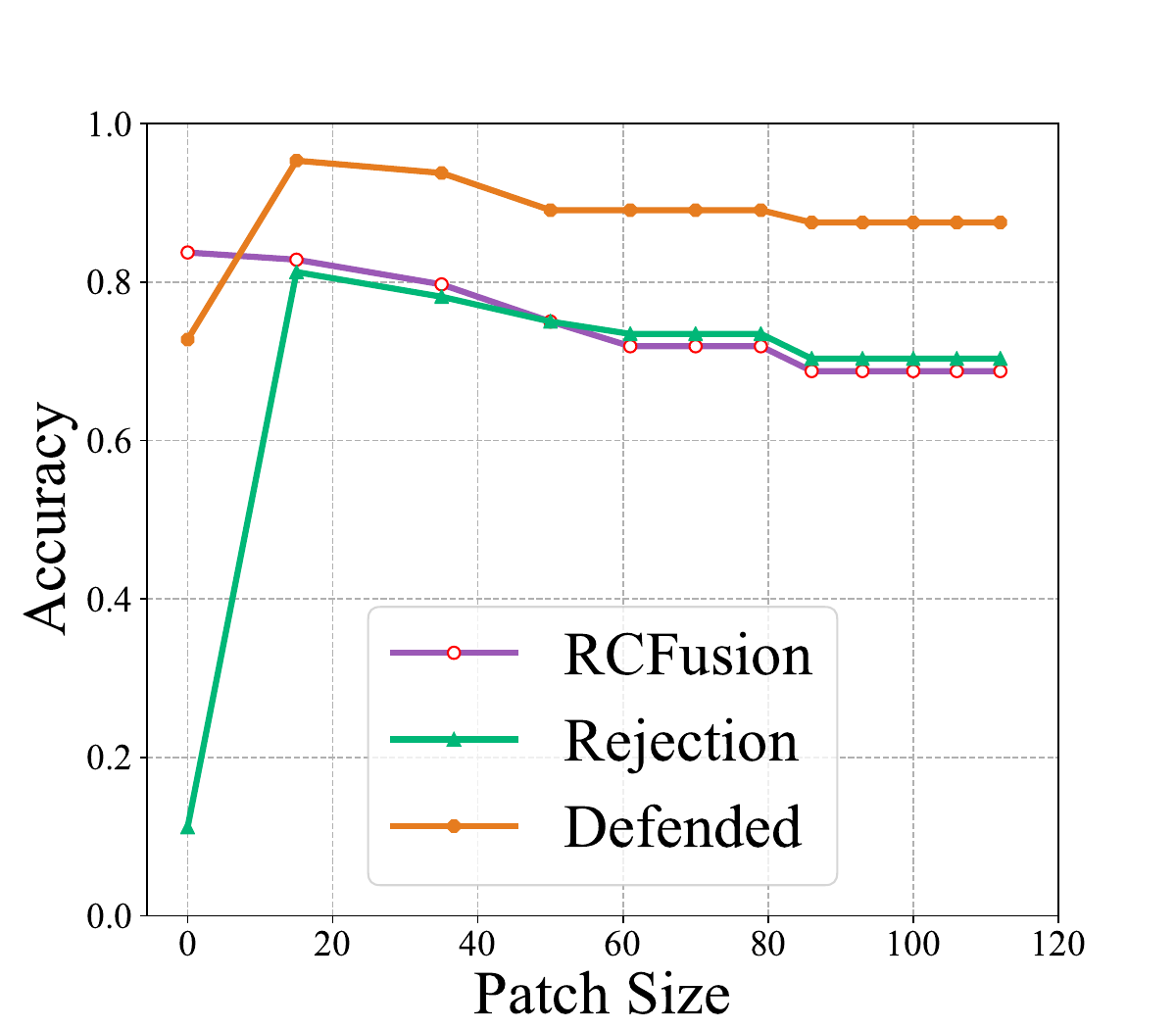}
    \includegraphics[width=0.48\textwidth]{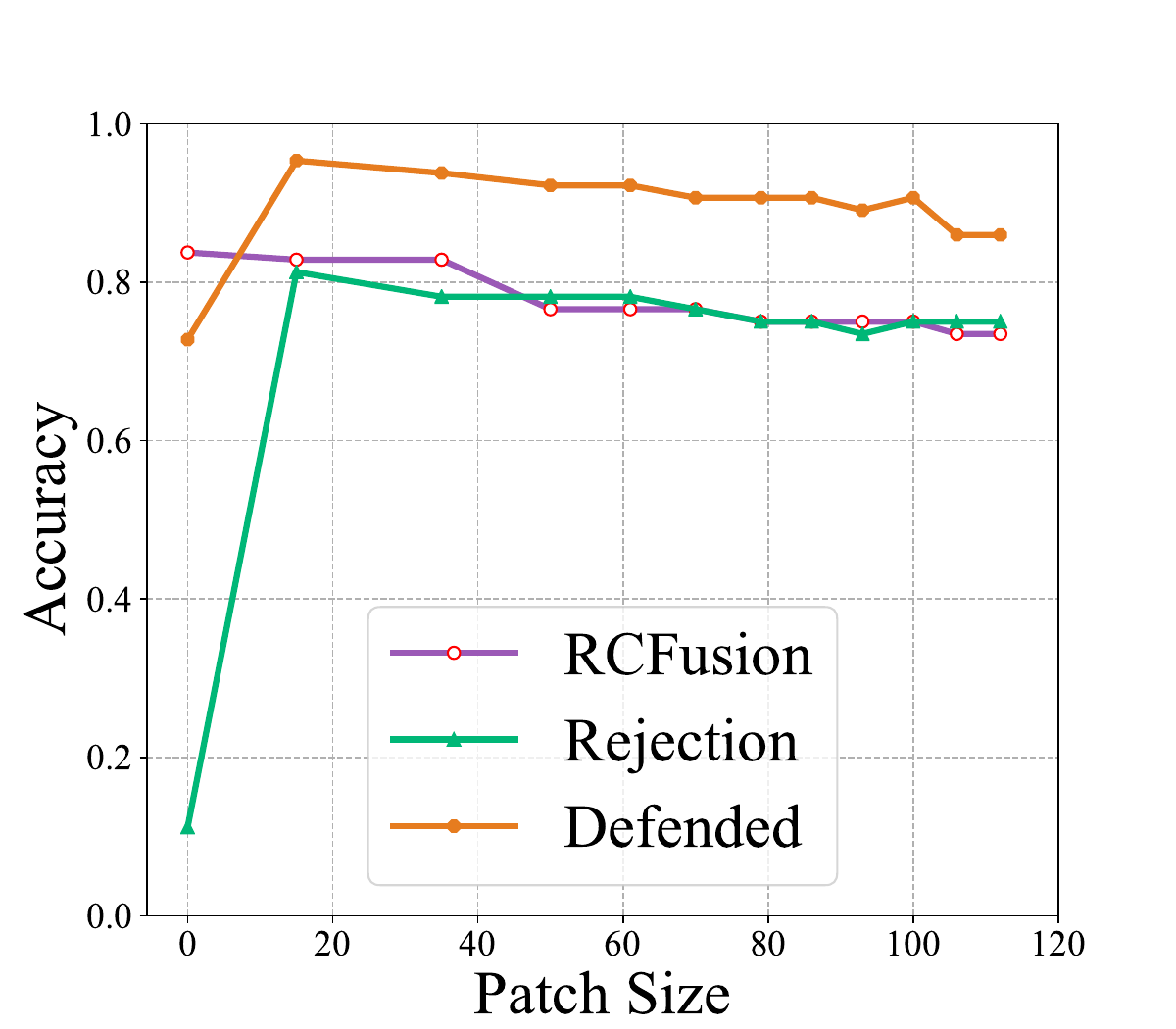}}

    {\includegraphics[width=0.48\textwidth]{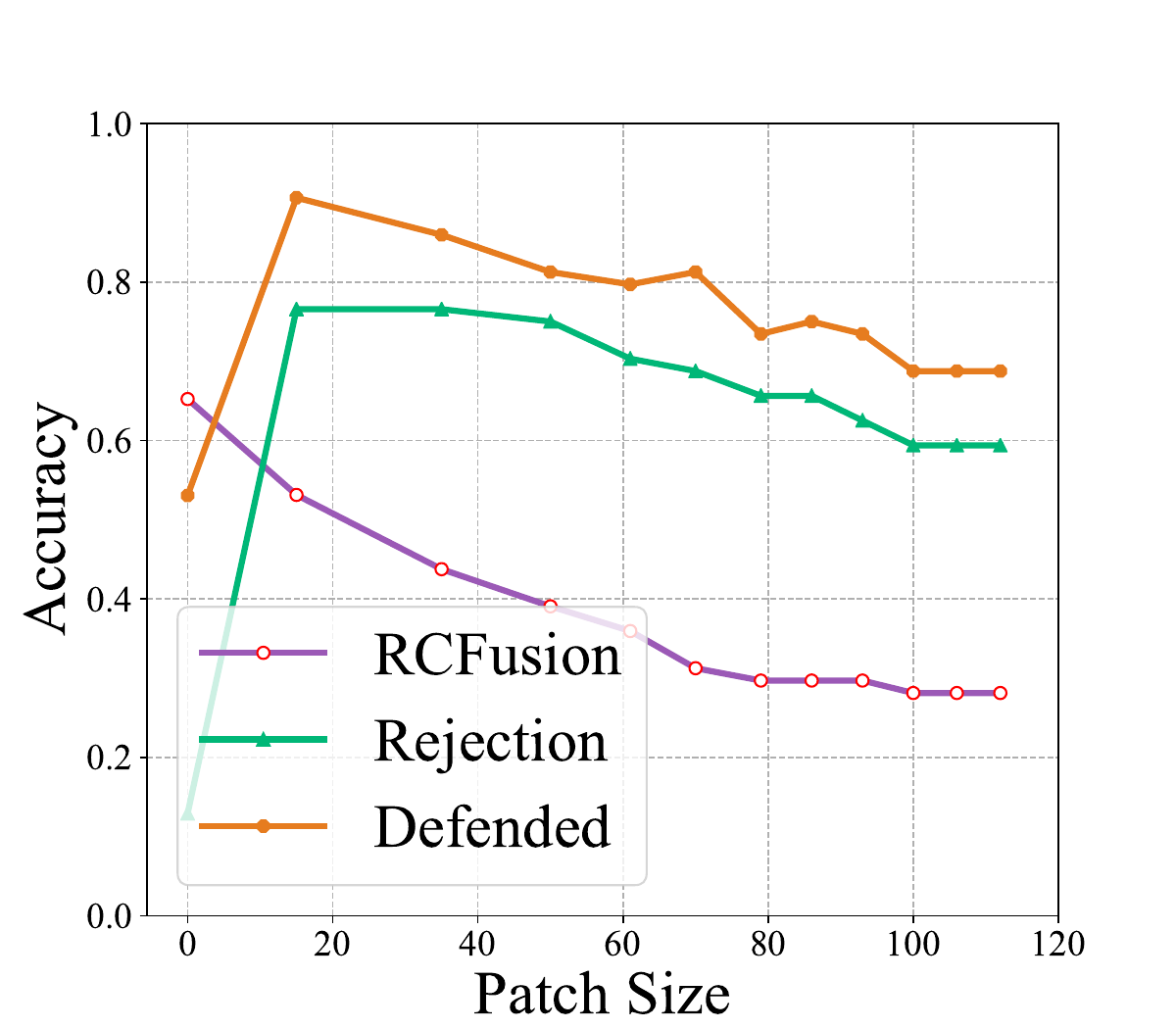}
    \includegraphics[width=0.48\textwidth]{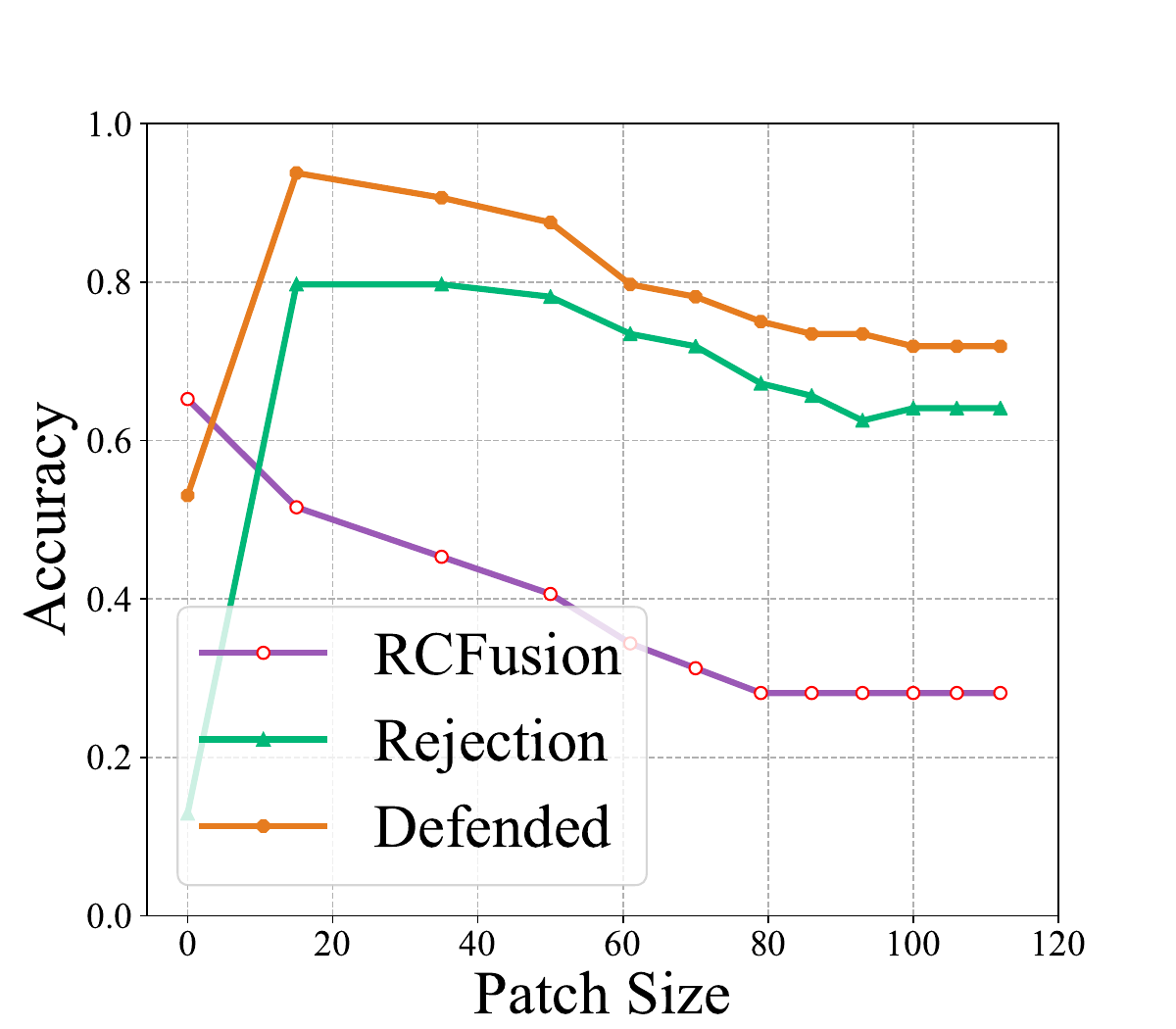}}
    \caption{\yzzedit{The robustness evaluation curve is computed by modifying only a portion image of the RGB-D (left) and RGB (right) channel of RCFusion constructed by AlexNet on the RGB-D Object Dataset (top-row) and the OCID Dataset (bottom-row).}}
\label{fig:rob_rcfusion_alexnet_patch}
\end{figure}

\myparagraph{Robustness of the Proposed Defenses.} We here inspect the robustness results offered by our rejection-based defensive method.
The results are reported in Fig.~\ref{fig:rob_resnet18}-~\ref{fig:rob_rcfusion_alexnet_patch}.

Firstly, we try to apply the rejection base mechanism on a simple ResNet-18 instead than to RCFusion. We report the result obtained when the attacker can modify the full input and the ResNet-18 is trained on the RGB (left column) and depth part of the input (right) for the two considered datasets. From Fig.~\ref{fig:rob_resnet18}, we can see that the accuracy of the defended ResNet-18 (called Defended in the figure) decreases fast when the classifier is under attack ($\epsilon > 0$). This means that applying the rejection mechanism to the ResNet-18 does not provide a notable advantage, except when the ResNet-18 is trained on the depth part of the RGB-D Object Dataset. 

\yzzedit{Therefore, we evaluate the defense's performance when applied to RCFusion constructed by ResNet-18. Also, to compare the impact of different backbone networks on RCFusion performance, we evaluate the performance of the defense when applied to RCFusion constructed by AlexNet.}
We present the results when the attacker can modify the full image in Fig.~\ref{fig:rob_rcfusion_all}-~\ref{fig:rob_rcfusion_alexnet_all}. We can see that the accuracy of defended classifier and RCFusion decreases until it reaches zero. This means that the attack algorithm we employ to perform the analysis works correctly. Moreover, from these two plots, we can see that even though the accuracy decreases when the perturbation increases, \textit{the accuracy of the defended RFCusion decreases more gracefully than that of the defended ResNet-18.}
\yzzedit{Besides, we also can see that the performance of the detector when applied to RCFusion constructed by ResNet-18 (Fig.~\ref{fig:rob_rcfusion_all}) drops faster than that of the detector when applied to RCFusion constructed by AlexNet (Fig.~\ref{fig:rob_rcfusion_alexnet_all}), in other words, the robustness of the detector when applied to RCFusion constructed by ResNet-18 is inferior to that of the detector when applied to RCFusion constructed by AlexNet.}

To explore the performance of the defended RCFusion in a more realistic scenario, we assess its robustness against adversarial patches. To this end, we perform the attack by modifying a portion of an image and present the result in Fig.~\ref{fig:rob_rcfusion_patch}-~\ref{fig:rob_rcfusion_alexnet_patch}. In the absence of attacks (i.e., \emph{patch size} $=$ 0), the performance of RCFusion slightly outperforms the classifier defended with the proposed detector. This is expected, and it is because a small portion of legitimate samples is incorrectly flagged as adversarial examples.
Under attack (\emph{patch size}  $>$ 0), the defended classifier shows more robustness than that of RCFusion, as the accuracy of the defended classifier decreases more gracefully than that of RCFusion when \emph{patch size} $>$ 0. It is worth noticing that the accuracy of the defended classifier even increases for a small \emph{patch size}, as the test samples immediately become blind-spot adversarial examples when modified slightly and end up in a region that is far from the rest of the data. Moreover, with the increase of the \emph{patch size}, the test samples gradually drifted inside a different class, making them indistinguishable for the rejection-based defense.
Overall, by comparing the performance of the defended classifier and the undefended RCFusion, and the rejection rate, we show that \textit{our defense mechanism provides a more robust performance than that of the undefended RCFusion under attack performed by modifying the full and a portion of an image}.

\myparagraph{Comparison with Adversarial Training.}
We compare the performance of our detector against the RCFusion model defended with the \emph{adversarial training} technique developed by Wang et al.~\citep{wang2022adversarial}.
The defense in \citep{wang2022adversarial} augments the training dataset with correctly labeled adversarial examples, thus helping the neural network better generalize when confronted with malicious noise.
To do so, we start by computing adversarial examples on the full images to attack the RGB-D part of RCFusion, with a perturbation $\epsilon = 0.1 (0.2)$ for RGB-D Object Dataset (OCID), and we use them along with the unperturbed training data to produce robust models (RCFusion\_AT).

\begin{figure}[t]
    \centering
    \includegraphics[trim=0 0 0 0, clip, width=.45\textwidth]{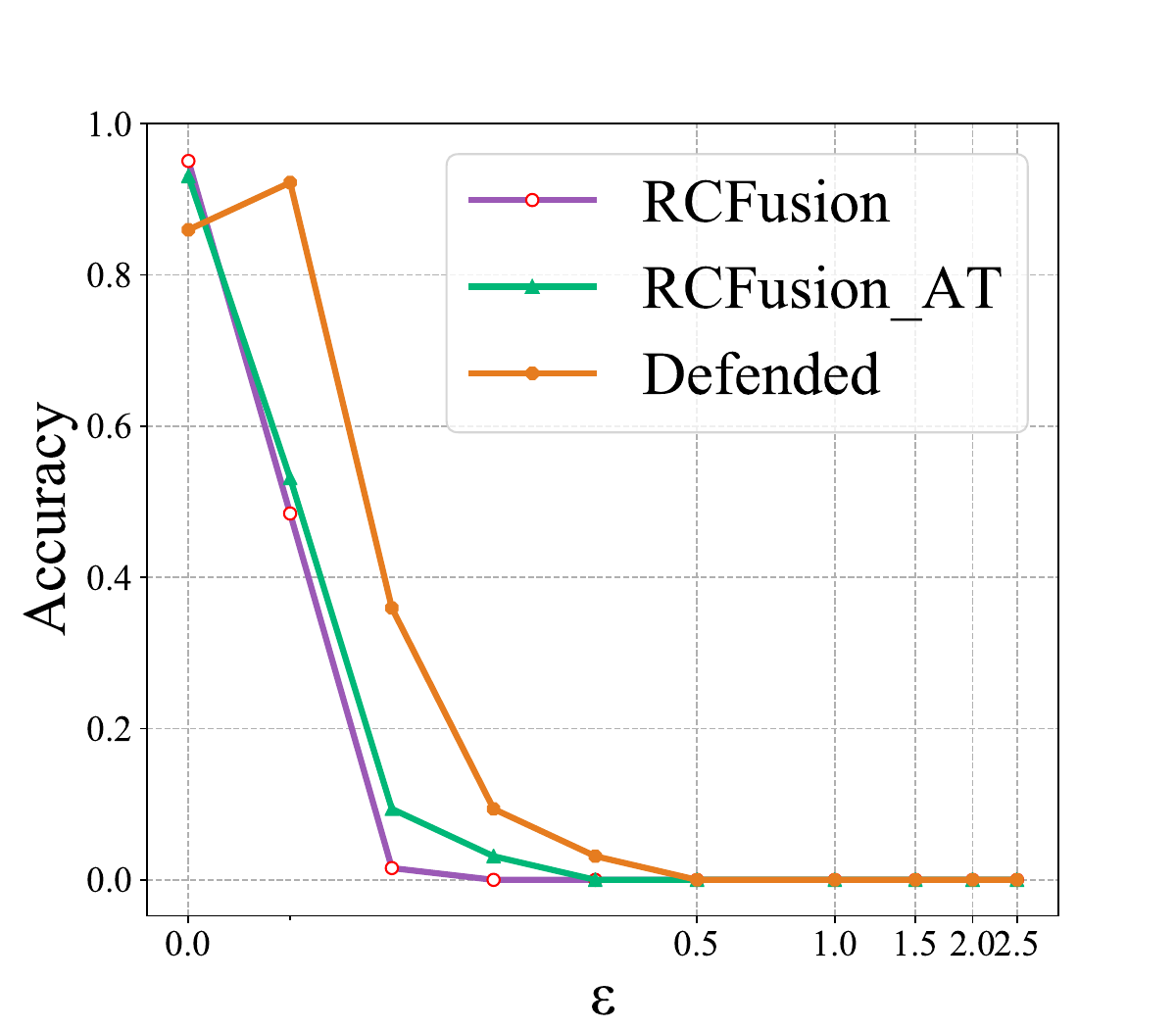}
	\includegraphics[trim=0 0 0 0, clip, width=.45\textwidth]{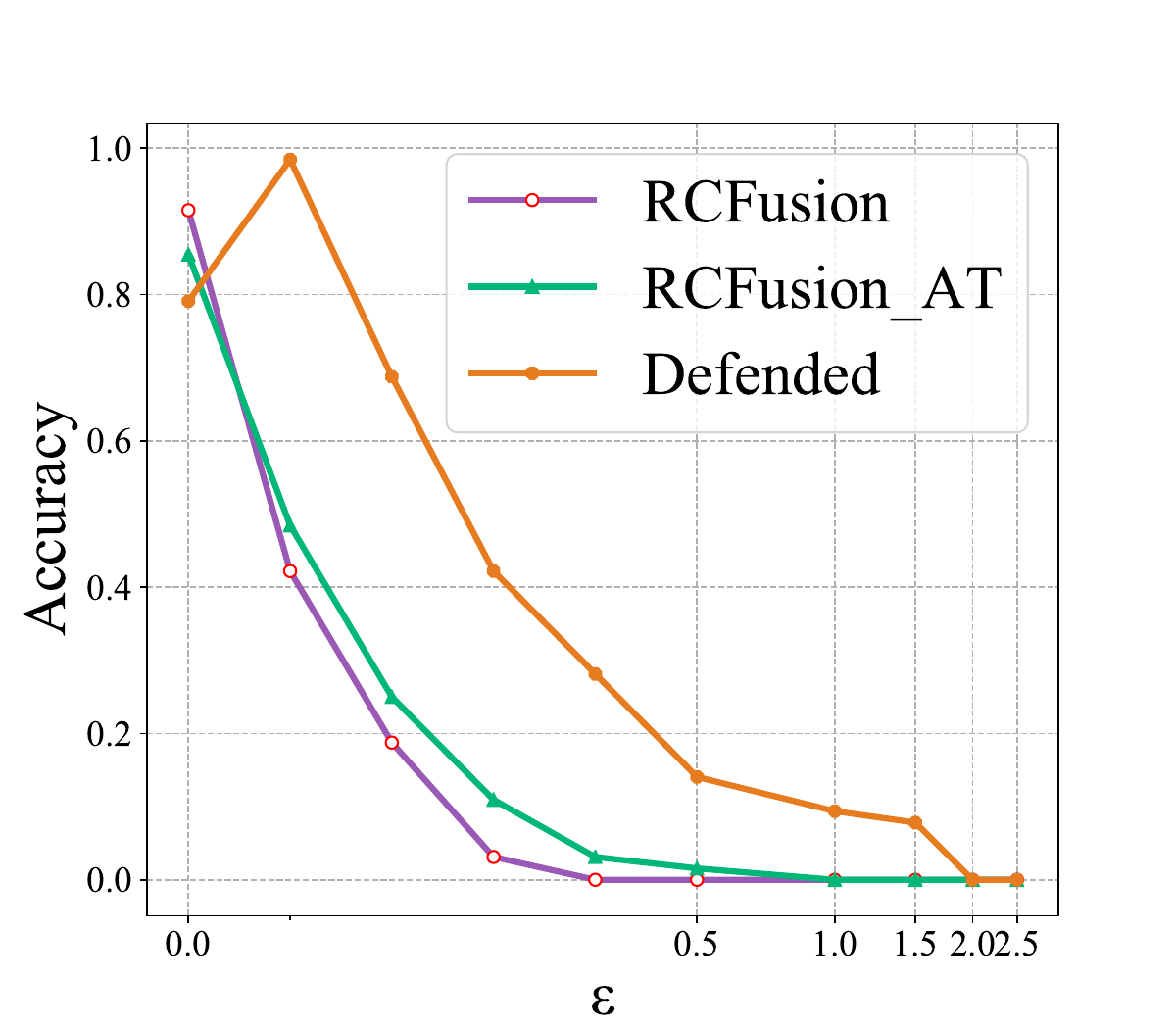}
    \caption{\yzzedit{Comparison between the robustness of the proposed defense methodology with the robustness of the adversarial training considering all considered $\epsilon$ on the RGB-D Object Dataset (left) and the OCID (right), respectively.}}
    \label{fig:adversarial_training_}
\end{figure}

\begin{figure}[t]
    \centering
    \includegraphics[trim=0 0 0 0, clip, width=.45\textwidth]{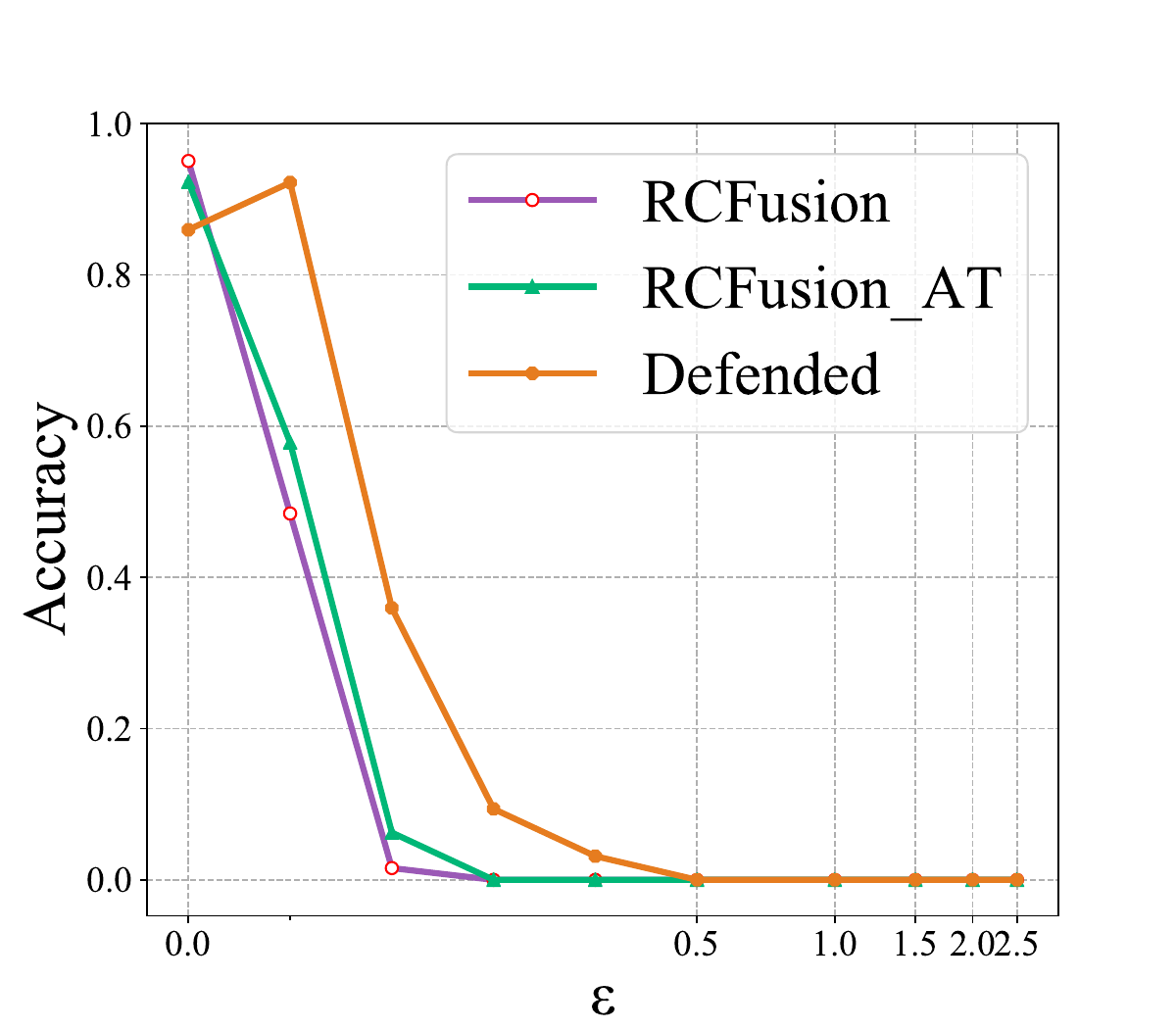}
	\includegraphics[trim=0 0 0 0, clip, width=.45\textwidth]{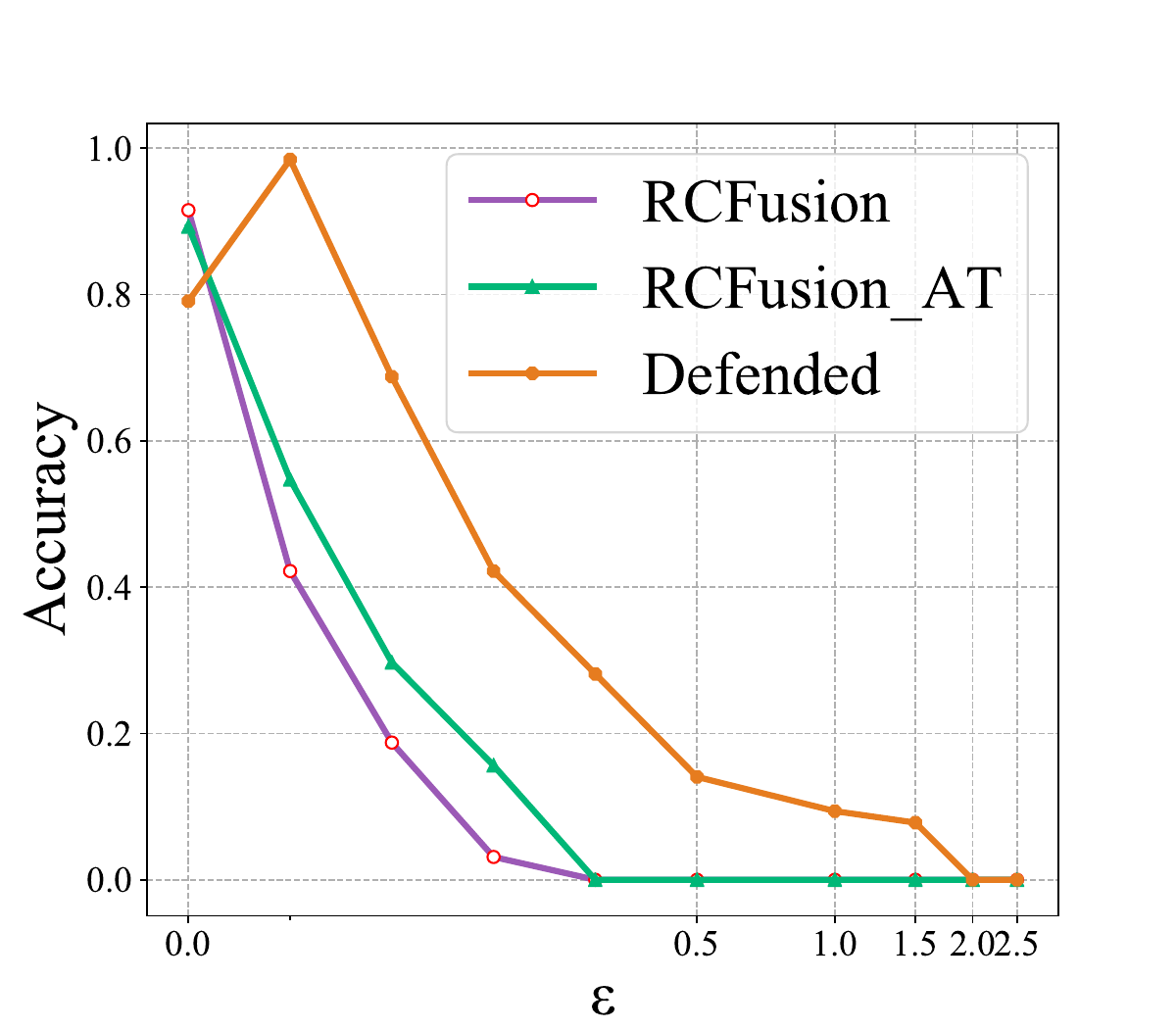}
    \caption{\yzzedit{Comparison between the robustness of the proposed defense methodology with the robustness of the adversarial training considering a single $\epsilon$ on the RGB-D Object Dataset (left) and the OCID (right), where $\epsilon=0.1 (0.2)$ for RGB-D Object Dataset (OCID).}}
    \label{fig:adversarial_training}
\end{figure}

 We depict the experimental comparison between our detector and the adversarial training defense by Wang et al.~\citep{wang2022adversarial} in Fig.\ref{fig:adversarial_training_} and Fig.\ref{fig:adversarial_training}.
\yzzedit{The key observation is that conventional adversarial training, for attacks injecting perturbations optimized by randomly-picking all considered $\epsilon$, \ie $\epsilon \in \{0.1,0.2,0.3,0.4,0.5,1,1.5,2,2.5\}$ (Fig.\ref{fig:adversarial_training_}), and by using an $\epsilon = 0.1 (0.2)$ for RGB-D Object Dataset (OCID) (Fig.\ref{fig:adversarial_training}) for RGB-D Object Dataset (OCID), provide only a small improvement in robustness.} In contrast, the proposed robustness provided by the proposed detector is more elevated. This is because, with adversarial training, as soon as the perturbation applied to the adversarial examples is slightly larger than the classifiers' margin, they are misclassified. Instead, using the proposed detector, they are classified as adversarial examples unless they become quite similar (in deep space) to the samples of the target class. Therefore, \textit{our defense provides higher robustness than adversarial training}. Moreover, they are complementary; therefore, they could be used jointly to obtain even more robustness.

In conclusion, our proposed technique demonstrates superior efficacy in countering adversarial attacks, enhancing model robustness against malicious perturbations while maintaining high accuracy and reliability in detecting out-of-distribution samples.


\clearpage

\clearpage

\clearpage

\clearpage

\myparagraph{Remarks.} In this work, we have assessed the performance of a state-of-the-art RGBD-based object recognition system called RCFusion against adversarial examples. Given that this system considers not only the RGB features but also the depth, it is reasonable to suppose it is more resilient to adversarial examples that change only the RGB part of the input compared to a system based only on RGB features. However, we have shown that their robustness is similar. Our results show that the vulnerability of RCFusion is mainly due to the usage of RGB features that, even if combined with the deep features, make the system vulnerable. However, they are necessary to obtain satisfactory performance. Therefore, we have proposed a defense based on a detection mechanism that, as we have shown, can make RCFusion more robust with negligible overhead. Moreover, we have shown that this defense mechanism is more effective than the only defenses proposed so far to secure RGBD-based systems~\citep{wang2022adversarial}. 

\section{Related Work}
\label{sec:related}
In the following, we discuss the work related to the vulnerability of RGB-D models and the previously proposed defenses against adversarial examples.\\

\myparagraph{Vulnerabilities of RGB-D models.} While it is straightforward to compute adversarial attacks against a machine learning model, understanding the rationale behind such weakness is a difficult task.

Geirhos et al.~\citep{GeirhosRMBWB19} remark on their discussion that humans most likely rely on the shape of observations to categorize and recognize objects, while deep neural networks retrieve information from the observed texture.
To support this intuition, the authors interview volunteers by asking them to classify silhouettes and textures of objects and feed the same inputs to RGB neural networks.
The authors also test the robustness of both human volunteers and neural networks by testing them with common corruptions applied to images, but they did not test adversarial attacks that target the shape or depth information.

Tu et al.~\citep{tu2021exploring} analyze the robustness of object detectors of self-driving cars that recognize objects by acquiring RGB images and proximity scans with Lidar sensors. 
The authors develop attacks against both components, jointly or separately, and show how much they degrade the performance of the target classifier.

Abdelfattah et al.~\citep{abdelfattah2021adversarial} similarly evaluate the robustness of RGB-D models against adversarial perturbations with the intent of misleading the cloud point reconstruction. They achieve this result by virtually creating a single object with an adversarial shape and texture.

Yu et al.~\citep{yu2020investigating} investigates the robustness of fusion models that leverage RGB and thermal information to compute image segmentation. 
Their results highlight that these models are not effective against adversaries, even if the attacks are conducted against one single part at a time.

Xie et al.~\citep{Xie2023Camera3D} investigate the adversarial robustness of 3D object recognition by considering a set of attacks, including pixel-based attacks, universal patches, and black-box attacks in the form of transferability attacks. Their main findings suggest that robust depth recognition can improve the adversarial robustness of RGB-D models.

Even if these recent works analyze the robustness of RGB-D models, they all lack an in-depth study regarding the reason for different levels of the robustness of the RGB and depth component.
Complementary to previous works, we empirically assess the robustness of both RGB and depth features, and we analyze the variability learned at training time in each internal layer to explain the reason behind depth robustness and RGB vulnerability. \\

\myparagraph{Adversarial defenses.}
So far, no work has studied the effectiveness of defenses on RGB-D object recognition systems. The only work that proposes a defense for an RGB-D system is the one proposed by Wang et al.~\citep{wang2022adversarial} that aims to secure an object detector. In this work, the authors study the application of adversarial training~\citep{madry2018towards} on both the RGB and depth components of fusion networks, and they discover that both accuracy and robustness decreased when hardening separately and jointly the two parts.

Many works, instead, have previously proposed defenses to secure RGB systems against adversarial examples. 

Crecchi, Sotgiu, et al.~\citep{sotgiu2020deep, crecchi2022fader} propose a detection mechanism that trains a machine learning model on the internal representation learned by the network to defend.
At test time, the detector discards all the input samples whose internal representation mismatches the one learned at training time.

Meng et al.~\citep{meng2017magnet} propose Magnet, a detector that intercepts anomalous samples by computing the difference between the input and its de-noised version, leveraging an autoencoder neural network.

All these methods leverage a detection mechanism similar to our proposed defense but not applied to RGB-D systems to discard adversarial patch attacks.

\section{Conclusions}
\label{sec:conclusions}
In this work, we investigate the lack of robustness of RGB-D systems by explaining that attackers can easily obtain misclassification thanks to the weakness introduced by the color information.
We explain this phenomenon by leveraging the Central Kernel Alignment metric, showing that models trained on RGB or both RGB and depth are more sensitive to minimal changes of input samples compared to networks trained only on depth, hence amplifying the weakness to adversarial examples.
To reduce the vulnerability of RGB-D systems, we develop a detector capable of discarding anomalous input samples by comparing their deep fusion representation with centroids computed at training time.
We empirically show that our defense mechanism can reduce the effect of adversarial examples and adversarial patches aimed to circumvent such detectors. Moreover, we have shown that the only approach that was proposed by previous works to defend RGBD-based systems, namely, adversarial training, can only slightly increase the robustness of RCFusion against adversarial examples with respect to the undefended model. Whereas the proposed approach is more effective in spite of being also less expensive at training time, as discussed in Section~\ref{sec:defense}. 

One limitation of our work is that our detector still uses the RGB information. It would be ideal to leverage only the depth channel of test samples since we showed that depth alone is more robust to minimal perturbation.
However, this may not be possible because the accuracy of the systems based on depth is much lower than those based on color.
Hence, in our future work, we will work on creating a detector based only on depth to obtain a more robust system and thus increase the perturbation that the attackers should apply to images to subvert the system.

\section*{Acknowledgments}
This work was partly supported by the PRIN 2017 project RexLearn, funded by the Italian Ministry of Education, University and Research (grant no. 2017TWNMH2); by BMK, BMDW, and the Province of Upper Austria in the frame of the COMET Programme managed by FFG in the COMET Module S3AI; by Spoke 10 ``Logistics and Freight'' within the Italian PNRR National Centre for Sustainable Mobility (MOST), CUP I53C22000720001; and by the Key Research and Development Program of Shaanxi (Program Nos. 2022ZDLGY06-07, 2021ZDLGY15-01, 2021ZDLGY09-04 and 2021GY-004), the International Science and Technology Cooperation Research Project of Shenzhen (GJHZ20200731095204013), the National Natural Science Foundation of China (Grant No. 61772419).

\bibliographystyle{elsarticle-num}
\bibliography{addref}

\newpage

\appendix
\section{Notation}~\label{appendix:notation}
\noindent We list the notation and symbols used throughout this paper in Table~\ref{tab:glossary}.

\begin{table}[h!]
\centering
 \caption{Notation and symbols used in the paper.}\label{tab:glossary}
  \renewcommand{\arraystretch}{1.2}
  \setlength\tabcolsep{2pt} 
  
   \resizebox{\textwidth}{!}{  
\begin{tabular}{@{}llll@{}}
\toprule
\multicolumn{4}{c}{\textbf{Adversarial Examples}}                                                                                                     \\ \midrule
$\vct x$              & input sample                                  & $y \in \mathcal{Y}$                  & true label of $\vct x$                        \\
$\vct \delta$         & adversarial perturbation                      & $\parallel \vct \delta \parallel_p $ & $\ell_p$-norm of $\vct \delta$         \\
$\epsilon$            & maximum noise magnitude                       & $\eta$                               & attack step size                       \\
$\Pi$                 & projection operator                           & $\mathcal{L}$                        & adversarial loss                       \\ \midrule
\multicolumn{4}{c}{\textbf{Defending RGB-D Classifier}}                                                                                               \\ \midrule
$\mathcal{S}(\vct x) \in \mathbb{R}^{1 \times c}$ &
  softmax output layer for input $\vct x$ &
  $s_i(\vct x)$ &
  logit score of class $i$ for sample $\vct x$ \\
$\mathcal{R}(\vct x)$ & RGB-D features of RCFusion for input $\vct x$ & $\gamma$                             & the predicted label of RCFusion        \\
$\tilde{\vct x}_k$ &
  $k$-th training sample &
  $\mathcal{C}_\gamma(\tilde{\vct x}_k)$ &
  centroid for the $\gamma$-th class \\
$n_{\gamma}$ &
  number of samples that belonging to class-$\gamma$ &
  $\mathcal{E}(\vct x)$ &
  anomaly score for input $\vct x$ \\
$\mathcal{S}'(\vct x)$ &
  softmax output layer for input $\vct x$ with rejection class &
  $\mathcal{L}_d$ &
  adaptive adversarial loss \\
$\beta$               & rejection threshold                           & $\rho$                               & $\beta$ optimal value search step size \\ \bottomrule
\end{tabular}
}
\end{table}

\vfill

\end{document}